\documentclass[preprint,12pt,authoryear]{elsarticle}
\usepackage{amssymb}
\usepackage{kotex}
\usepackage[normalem]{ulem}

\usepackage{pifont}
\usepackage{hyperref}
\usepackage{amsmath}
\usepackage{booktabs}
\usepackage{amsmath}
\usepackage{graphicx}
\usepackage{hhline}
\usepackage{tabularray}
\usepackage{appendix}
\usepackage{array}
\usepackage{multirow}
\usepackage[normalem]{ulem}
\usepackage{graphicx}
\usepackage{multirow}
\usepackage{colortbl}
\usepackage{booktabs}
\usepackage{adjustbox}
\usepackage{xcolor}
\usepackage{tabularx}
\usepackage{longtable}
\usepackage{caption}
\usepackage{microtype}
\useunder{\uline}{\ul}{}

\definecolor{lightblue}{RGB}{51,153,255}
\definecolor{lightred}{RGB}{255,102,102}
\hypersetup{colorlinks = true, linkcolor = blue, anchorcolor = red, citecolor = blue, filecolor = red, urlcolor = red,
            pdfauthor=author}

\journal{Computer Vision and Image Understanding}

\begin{document}

\begin{frontmatter}
\title{Domain Adaptation of Attention Heads for Zero-shot Anomaly Detection}

\author[1]{Kiyoon Jeong\fnref{fn1}}
\ead{kiyoon_jeong@korea.ac.kr}
\author[2]{Jaehyuk Heo\fnref{fn1}}
\ead{jaehyuk.heo@snu.ac.kr}
\author[1]{Junyeong Son}
\ead{junyeong_son@korea.ac.kr}
\author[2]{Pilsung Kang\corref{cor1}}
\ead{pilsung\_kang@snu.ac.kr}

\fntext[fn1]{These authors contributed equally to this work.}

\affiliation[1]{organization={Department of Industrial and Management Engineering, Korea University},
    addressline={145, Anam-Ro, Seongbuk-Gu}, 
    city={Seoul},
    postcode={02845},
    country={Republic of Korea}}
\affiliation[2]{organization={Department of Industrial Engineering, Seoul National University},
    addressline={1, Gwanak-ro, Gwanak-gu}, 
    city={Seoul},
    postcode={08826},
    country={Republic of Korea}} 
\cortext[cor1]{Corresponding author. Tel.: +82-2-880-7360}

\begin{abstract}
Zero-shot anomaly detection (ZSAD) enables anomaly detection without normal samples from target categories, addressing scenarios where task-specific training data is unavailable. However, existing ZSAD methods either neglect adaptation of vision-language models to anomaly detection or implement only partial adaptation. This paper proposes Head-adaptive CLIP (HeadCLIP), which effectively adapts both text and image encoders. HeadCLIP employs learnable prompts in the text encoder to generalize normality and abnormality concepts, and introduces \textit{learnable head weights} in the image encoder to dynamically adjust attention head features for task-specific adaptation. A \textit{joint anomaly score} is further proposed to leverage adapted pixel-level information for enhanced image-level detection. Experiments on 17 datasets across industrial and medical domains demonstrate that HeadCLIP outperforms existing ZSAD methods at both pixel and image levels, achieving improvements of up to 4.9\%p in pixel-level mean anomaly detection score (mAD) and 3.7\%p in image-level mAD in the industrial domain, with comparable gains (3.2\%p, 3.2\%p) in the medical domain. Code and pretrained weights are available at \url{https://github.com/kiyoonjeong0305/HeadCLIP}.
\end{abstract}

\begin{keyword}
Image Anomaly Detection \sep Zero-shot Anomaly Detection \sep Vision-Language Model
\end{keyword}

\end{frontmatter}

\section{Introduction}\label{Introduction}

Visual anomaly detection is essential for manufacturing inspection \citep{xie2024iad}, financial security \citep{Park2024EnhancingAD}, and medical diagnosis \citep{kumari2023comprehensive}. Unsupervised anomaly detection (UAD), which learns patterns solely from normal data \citep{deepIAD}, has been widely adopted when anomalies are rare or diverse. However, UAD suffers from the cold start problem \citep{adaclip}: normal data may be unavailable during early deployment of new manufacturing processes \citep{adaclip}, or inaccessible due to privacy constraints in healthcare \citep{xie2024iad}.

To overcome this limitation, zero-shot anomaly detection (ZSAD) has emerged as a promising alternative that eliminates the need for target-category training data. Following the standard ZSAD protocol, models are trained on auxiliary datasets (e.g., MVTec AD) with full supervision to learn generalizable concepts of normality and abnormality, then evaluated on entirely unseen target categories (e.g., VisA, medical datasets). This cross-dataset transfer eliminates the need for category-specific data collection and annotation.

Central to recent ZSAD advances is CLIP \citep{clip}, a vision-language model pre-trained on hundreds of millions of web image-text pairs \citep{laion}. ZSAD methods leverage CLIP's generalization capability by computing similarity between visual features and textual prompts describing normal or abnormal states \citep{anovl, winclip, anomalyclip, adaclip}. However, since CLIP is pre-trained primarily on natural images, its effectiveness degrades in specialized domains such as industrial inspection and medical imaging, where visual statistics differ substantially from natural scenes.

\begin{figure}[t!]
    \centering
    \includegraphics[width=1\linewidth]{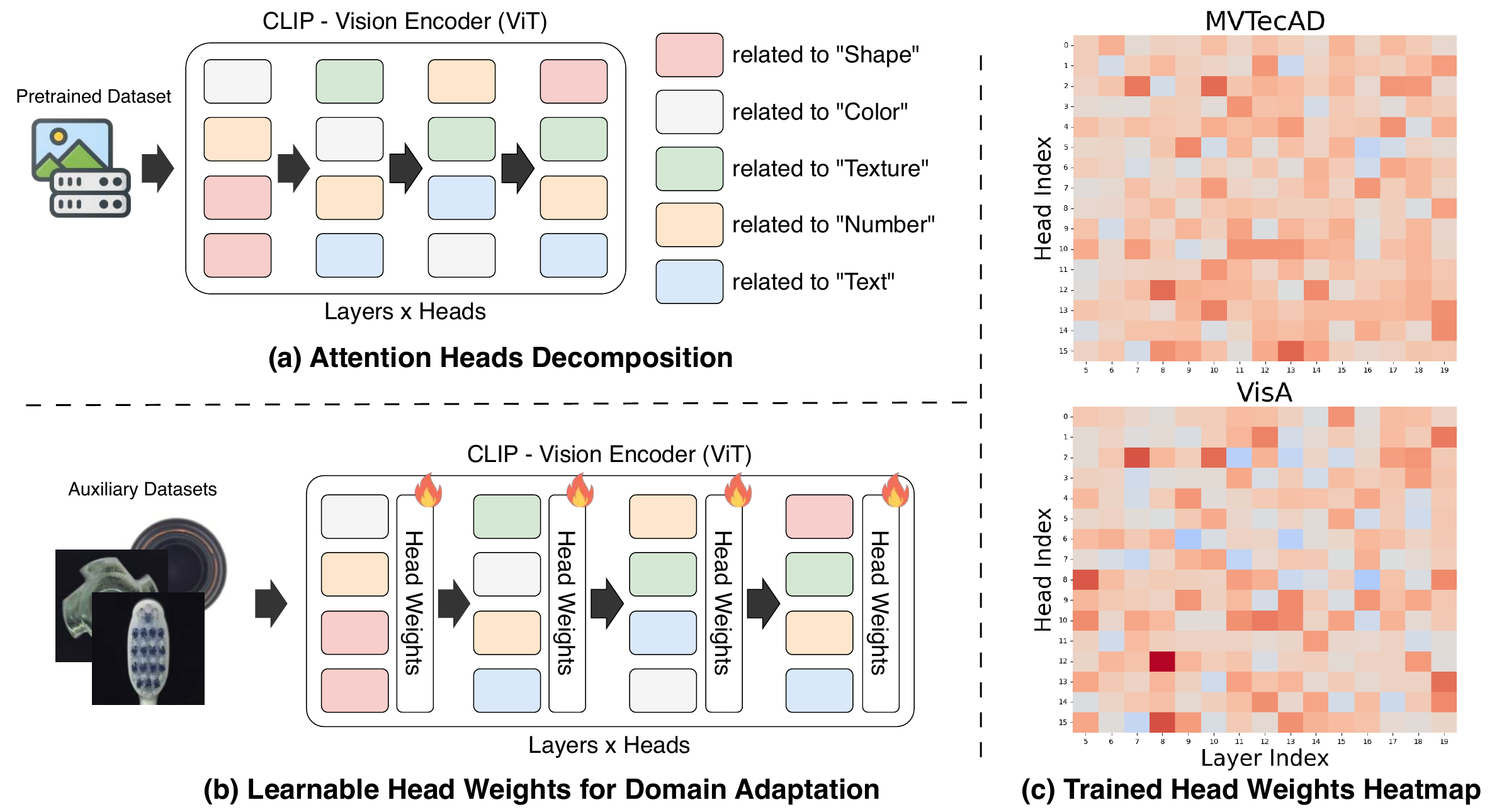}
    \caption{Overview of domain adaptation approaches in CLIP-based zero-shot anomaly detection. (a) Attention heads in CLIP's ViT capture distinct visual semantics~\citep{textspan}. (b) HeadCLIP introduces learnable head weights (indicated by fire icons) trained on auxiliary datasets to adapt the image encoder for industrial anomaly detection. (c) Trained head weight heatmaps on MVTecAD and VisA datasets, where red indicates high importance and blue indicates low importance.}
    \label{fig:intro}
\end{figure}

To bridge this domain gap, existing methods have explored various adaptation strategies, yet they remain incomplete. AnoVL~\citep{anovl} and WinCLIP~\citep{winclip} rely on handcrafted prompts without modifying CLIP's encoders. AnomalyCLIP~\citep{anomalyclip} and AdaCLIP~\citep{adaclip} fine-tune learnable prompts but leave the image encoder frozen. Critically, none of these methods exploit the internal structure of CLIP's Vision Transformer (ViT), where different attention heads capture distinct visual semantics such as shape, color, and texture~\citep{textspan} (Fig.~\ref{fig:intro}a).

To address these limitations, this paper proposes Head-adaptive CLIP (HeadCLIP), a ZSAD framework that enables effective adaptation of both text and image encoders (Fig.~\ref{fig:intro}). HeadCLIP employs learnable prompt tokens for generalizable normal/abnormal concepts and introduces learnable head weights (LHW) that selectively weight attention heads based on their task relevance (Fig.~\ref{fig:intro}b). The learned weight patterns show remarkable consistency across industrial datasets (Fig.~\ref{fig:intro}c), demonstrating that HeadCLIP identifies domain-characteristic features rather than dataset-specific artifacts. Additionally, a joint anomaly score (JAS) integrates pixel-level and image-level cues for robust detection.

Through these contributions, HeadCLIP achieves strong ZSAD performance in settings where normal data is scarce or inaccessible. Experiments across 17 datasets from industrial and medical domains demonstrate consistent improvements over prior ZSAD methods in both classification and segmentation tasks. The main contributions are summarized as follows:

\begin{itemize}
\item A novel domain adaptation method for CLIP's image encoder through learnable attention head weights in ViTs, exploiting head-specific semantic focus for enhanced anomaly detection.
\item A joint scoring strategy integrating domain-adapted pixel-level and image-level anomaly cues, yielding robust results across diverse anomaly types.
\end{itemize}

The remainder of this paper is organized as follows. Section 2 reviews related work. Section 3 details the HeadCLIP framework. Section 4 presents experiments and analysis. Section 5 concludes with future directions.

\section{Related Works}\label{Related works}

\subsection{Anomaly Detection}
Anomaly detection identifies patterns that deviate significantly from expected distributions. In industrial settings, it enables automated quality control by detecting visual defects \citep{yi2020patch, rudolph2023asymmetric, hyun2024reconpatch}, while in medical imaging, it supports early detection of pathological irregularities \citep{zhao2021anomaly, wang2021weakly, kascenas2022denoising}. The core challenge lies in the ill-defined nature of anomalies: they are rare, diverse, and lack consistent definitions \citep{zhu2024anomaly}, rendering supervised approaches impractical due to the cost of comprehensive annotation \citep{chiu2023self}.

\subsection{Data-Efficient Anomaly Detection}
UAD methods learn solely from normal samples and flag deviations at test time. These approaches fall into two categories. Density-based methods model normal data distributions through normalizing flows \citep{csflow, fastflow, yao2024local} or memory banks storing representative features \citep{padim, yi2020patch, guo2023recontrast}. Reconstruction-based methods assume that models trained on normal data will fail to reconstruct anomalies accurately, with architectures evolving from autoencoders \citep{draem, easynet, guo2023encoder, lu2024anomaly} and GANs \citep{scadn, dfmgan} to Transformers \citep{uniad, fod} and diffusion models \citep{diffusionad, ddad}.

Traditional UAD methods train separate models for each product category, limiting scalability in real-world deployments. To address this, multi-class anomaly detection has emerged as an active research direction, aiming to train a unified model across multiple categories \citep{yao2024prior, guo2025dinomaly}. While these approaches improve deployment efficiency, they still require sufficient normal samples from all target categories during training.

To further reduce data requirements, few-shot anomaly detection methods have been proposed that leverage only a small number of normal samples from target categories \citep{luo2025exploring}. However, these approaches still assume access to target domain data, which may be unavailable in scenarios involving new product lines or privacy-sensitive domains.

\begin{table}[!t]
\caption{Comparison of CLIP-based zero-shot anomaly detection methods. The table compares key design choices including text template type, text encoder adaptation strategy, vision encoder adaptation, auxiliary dataset requirement, and the number of learnable parameters. HeadCLIP achieves comprehensive adaptation of both encoders while maintaining parameter efficiency through learnable head weights, in contrast to prior methods that either leave encoders frozen or introduce heavyweight adapter layers.} \label{tab:rel_works}
\vspace{5pt}
\centering
\small
\begin{adjustbox}{max width=\textwidth}
\begin{tabular}{@{}lcccccc@{}}
\toprule
Models &
  \begin{tabular}[c]{@{}c@{}}Auxiliary\\      Dataset\end{tabular} &
  \begin{tabular}[c]{@{}c@{}}Text\\      Template\end{tabular} &
  Text Encoder &
  \# Text Params &
  Visual Encoder &
  \# Visual Params \\ \midrule
\begin{tabular}[c]{@{}l@{}}WinCLIP\\\citep{winclip}\end{tabular} &
  X &
  \begin{tabular}[c]{@{}c@{}}Class\\-specific\end{tabular} &
  Hand-crafted &
  0 &
  X &
  0 \\ \midrule 
\begin{tabular}[c]{@{}l@{}}AnoCLIP\\\citep{anovl}\end{tabular} &
  X &
  \begin{tabular}[c]{@{}c@{}}Class\\-specific\end{tabular} &
  Hand-crafted &
  0 &
  X &
  0 \\ \midrule
\begin{tabular}[c]{@{}l@{}}AnomalyCLIP\\\citep{anomalyclip}\end{tabular} &
  O &
  \begin{tabular}[c]{@{}c@{}}Class\\-agnostic\end{tabular} &
  Learnable Prompts &
  5,655,194 &
  X &
  0 \\ \midrule
\begin{tabular}[c]{@{}l@{}}AdaCLIP\\\citep{adaclip}\end{tabular} &
  O &
  \begin{tabular}[c]{@{}c@{}}Class\\-specific\end{tabular} &
  \begin{tabular}[c]{@{}c@{}}Hand-crafted\\+Learnable Prompts\end{tabular} &
  6,707,712 &
  Learnable Prompts &
  3,957,760 \\ \midrule
\begin{tabular}[c]{@{}l@{}}AA-CLIP\\\citep{ma2025aa}\end{tabular} &
  O &
  \begin{tabular}[c]{@{}c@{}}Class\\-specific\end{tabular} &
  \begin{tabular}[c]{@{}c@{}}Hand-crafted\\+Adapter Layers\end{tabular} &
  2,359,296 &
  Adapter Layers &
  10,223,616 \\ \midrule
\begin{tabular}[c]{@{}l@{}}AF-CLIP\\\citep{fang2025af}\end{tabular} &
  O &
  \begin{tabular}[c]{@{}c@{}}Class\\-agnostic\end{tabular} &
  Learnable Prompts &
  9,216 &
  Adapter Layers &
  2,100,224 \\ \midrule
\begin{tabular}[c]{@{}l@{}}HeadCLIP\\(Ours)\end{tabular} &
  O &
  \begin{tabular}[c]{@{}c@{}}Class\\-agnostic\end{tabular} &
  Learnable Prompts &
  5,655,194 &
  Learnable Head Weights &
  304 \\ \bottomrule
\end{tabular}
\end{adjustbox}
\end{table}

These limitations motivate ZSAD, which eliminates the need for any target-category training data by leveraging pre-trained vision-language models.

\subsection{Zero-shot Anomaly Detection}
ZSAD addresses the data accessibility challenge by leveraging pre-trained VLMs, particularly CLIP \citep{clip}, to detect anomalies without target-category training data. As summarized in Table~\ref{tab:rel_works}, CLIP-based ZSAD methods differ in their adaptation strategies.

Early approaches such as WinCLIP~\citep{winclip} and AnoVL~\citep{anovl} employed class-specific handcrafted prompts without modifying CLIP's encoders. While effective for injecting domain knowledge, these methods rely heavily on manual prompt engineering and generalize poorly beyond CLIP's training distribution.

AnomalyCLIP~\citep{anomalyclip} addressed this by introducing class-agnostic learnable prompts (5.7M parameters), enabling generalizable representations without category-specific templates. However, the frozen vision encoder limits performance in domains that differ significantly from CLIP's pretraining corpus.

Recent methods have explored dual-encoder adaptation. AdaCLIP~\citep{adaclip} combines handcrafted templates with learnable prompts for both encoders (6.7M + 4.0M parameters), though class-specific templates constrain generalization. AA-CLIP~\citep{ma2025aa} employs residual adapters (2.4M + 10M parameters), while AF-CLIP~\citep{fang2025af} introduces lightweight adapters with multi-scale aggregation (9K + 2.1M parameters).

In contrast to these adapter-based methods that introduce external modules, HeadCLIP proposes a fundamentally different strategy by leveraging the internal structure of ViTs. Based on the finding that different attention heads capture distinct visual semantics~\citep{textspan}, HeadCLIP learns attention head weights (304 parameters) to selectively emphasize task-relevant heads. Combined with class-agnostic learnable prompts (5.7M parameters), this internal reweighting achieves comprehensive encoder adaptation through minimal architectural modification.

\section{Proposed Method}

\subsection{Overview of HeadCLIP Framework}

This paper presents HeadCLIP, a vision-language-based anomaly detection framework that introduces learnable head weights to enhance fine-grained feature extraction in the context of industrial anomaly detection. As illustrated in Fig.~1(a), the framework builds upon the dual-path architecture introduced in prior work~\citep{clipsurgery, anomalyclip, adaclip}, which processes visual features through parallel global and local pathways using standard multi-head self-attention (MHSA) and consistent self-attention (CSA), respectively. 

The primary contributions of HeadCLIP are twofold: (1) the introduction of LHW within the multi-head consistent self-attention mechanism to adaptively emphasize discriminative attention heads, and (2) a JAS that effectively combines global semantic understanding with local spatial analysis. To train the learnable parameters efficiently, the proposed method employs auxiliary datasets with anomaly annotations, enabling the model to learn generalizable anomaly patterns that transfer to target industrial datasets in a zero-shot manner.

\begin{figure}[t!]
    \centering
    \includegraphics[width=1\linewidth]{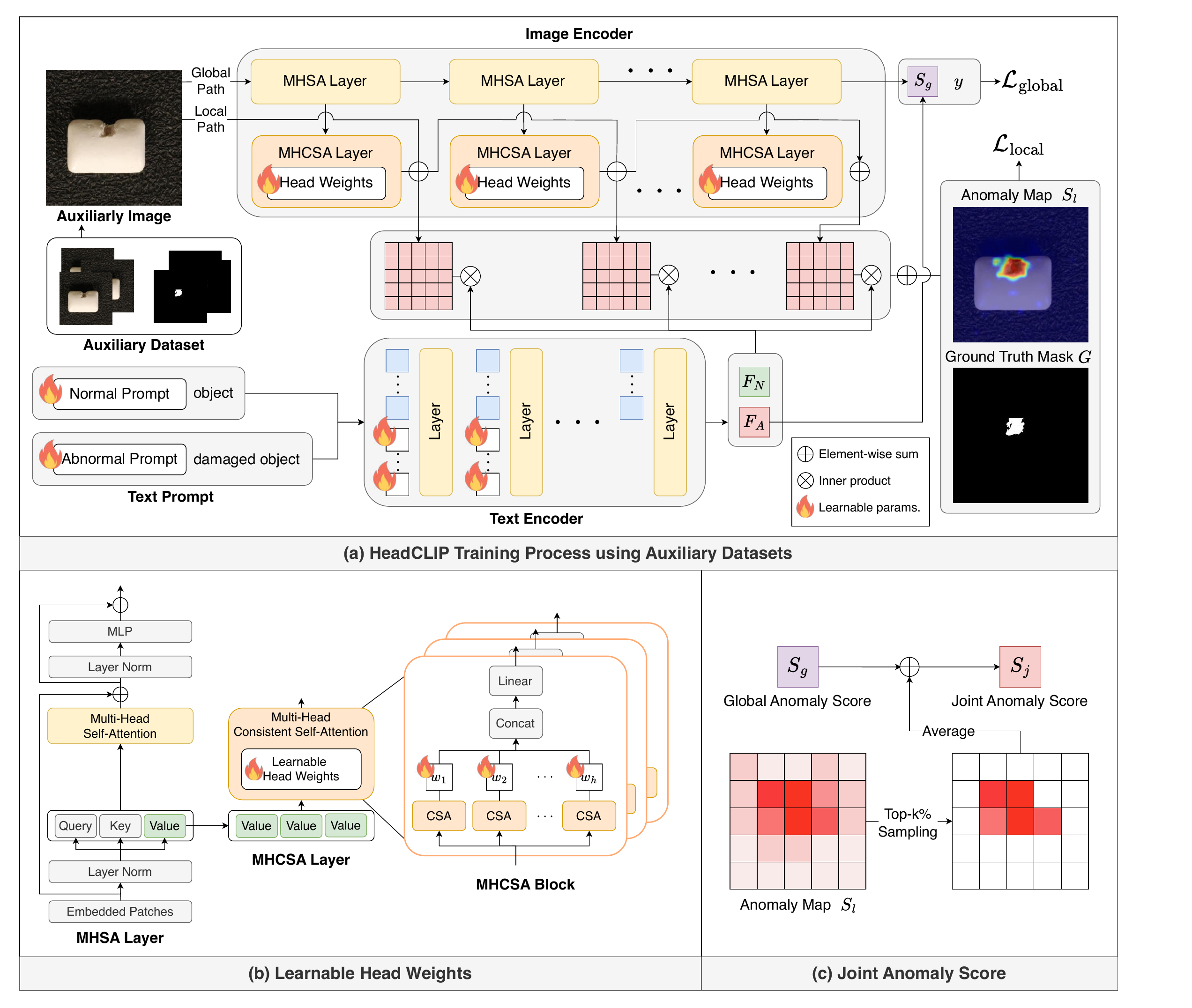}
    \caption{Overview of the HeadCLIP framework for anomaly detection. 
(a) Training process using auxiliary datasets. The image encoder consists of a global path with standard MHSA layers and a local path with MHCSA layers enhanced by LHW. The text encoder processes normal and abnormal prompts to provide semantic guidance. Auxiliary datasets are used to train the learnable parameters, generating anomaly maps that are supervised by ground truth masks through global loss $\mathcal{L}_{\text{global}}$ and local loss $\mathcal{L}_{\text{local}}$. 
(b) Detailed architecture of the MHCSA layer compared to the standard MHSA layer. The MHCSA block applies learnable weights ($w_1, w_2, \ldots, w_h$) to each CSA head, with outputs concatenated and projected through a linear layer. 
(c) Computation of the JAS by combining the global anomaly score $S_g$ with the top-$k\%$ average of the local anomaly map $S_l$, producing the joint anomaly score $S_j$.}\label{fig:main_figure}
\end{figure}

\subsection{Image Encoder with Dual-Path Architecture}

Following the architecture proposed in~\citep{anomalyclip}, the image encoder adopts a dual-path structure based on the ViT backbone, processing the input image $\mathbf{I} \in \mathbb{R}^{H \times W \times 3}$ through two parallel streams.

\textbf{Global Path.} The global path employs standard MHSA layers to extract holistic semantic features. Given an input image, it is first partitioned into non-overlapping patches of size $P \times P$ and linearly embedded into tokens. These tokens are then augmented with positional embeddings and processed through $L$ transformer layers with standard MHSA, producing global features $\mathbf{F}_g \in \mathbb{R}^{N \times D}$, where $N = HW/P^2$ denotes the number of patches and $D$ represents the feature dimension. The global path maintains the pretrained CLIP weights frozen, thereby preserving the rich semantic knowledge acquired from large-scale vision-language pretraining.

\textbf{Local Path.} The local path replaces standard MHSA with multi-head consistent self-attention (MHCSA). Starting from identical patch embeddings as the global path, the local path processes features through $L$ MHCSA layers, generating local features $\mathbf{F}_l \in \mathbb{R}^{N \times D}$. The outputs from both paths serve complementary purposes: the global features $\mathbf{F}_g$ contribute to holistic anomaly assessment, while the local features $\mathbf{F}_l$ provide pixel-level anomaly localization through detailed spatial analysis.

\subsection{Multi-Head Consistent Self-Attention with Learnable Head Weights}

The primary contribution of HeadCLIP lies in the multi-head consistent self-attention mechanism enhanced with learnable head weights. Recent studies have demonstrated that different attention heads in ViTs capture features with varying levels of discriminative importance~\citep{textspan}. Motivated by this finding, learnable head weights are introduced to adaptively modulate each attention head's contribution for industrial domain adaptation. This design addresses three key challenges: (1) not all attention heads in the pretrained CLIP encoder are equally relevant for industrial inspection tasks, (2) CLIP's pretraining on natural images (e.g., LAION) produces feature representations that are suboptimal for industrial domain characteristics, and (3) diverse anomaly patterns in industrial settings necessitate flexible feature extraction mechanisms. By learning to emphasize heads that extract domain-relevant features while suppressing those tuned for natural images, the proposed approach enables effective adaptation from natural to industrial domains without modifying the pretrained backbone.

\textbf{Consistent Self-Attention.} Unlike standard self-attention that computes attention weights using separate query and key projections with different parameters ($\phi_q$, $\phi_k$), CSA employs only the value projection parameter ($\phi_v$) to compute both the attention weights and output features. Specifically, for the $i$-th attention head, the CSA operation is defined as:

\begin{equation}
\text{CSA}_i(\mathbf{V}_i) = \text{softmax}\left(\frac{\mathbf{V}_i \mathbf{V}_i^T}{\sqrt{d_k}}\right) \mathbf{V}_i
\end{equation}

where $\mathbf{V}_i \in \mathbb{R}^{N \times d_k}$ denotes the value projection obtained from standard MHSA using parameter $\phi_v$, and $d_k = D/h$ represents the dimension per head with $h$ being the total number of heads. This design ensures that the self-attention matrix $\mathbf{V}_i \mathbf{V}_i^T$ is computed using homogeneous parameters, thereby building relations among tokens with consistent semantic regions~\citep{clipsurgery}. By utilizing only the value features, CSA enables the local path to capture fine-grained local semantic information more effectively, as the attention mechanism naturally emphasizes spatially proximate and semantically similar tokens.

\textbf{Learnable Head Weights.} As depicted in Fig.~1(b), learnable scalar weights $\{w_1, w_2, \ldots, w_h\}$ are introduced to modulate the contribution of each attention head for domain adaptation. The weighted multi-head CSA is computed as:

\begin{equation}
\text{MHCSA}(\mathbf{Q}, \mathbf{K}, \mathbf{V}) = \text{Linear}\left(\text{Concat}(w_1 \cdot \text{CSA}_1, \ldots, w_h \cdot \text{CSA}_h)\right)
\end{equation}

where the learnable weights $\{w_i\}_{i=1}^h$ are optimized during training to emphasize heads that extract features relevant to industrial domains while down-weighting heads specialized for natural image characteristics. This mechanism allows the model to automatically discover which attention patterns are most suitable for industrial inspection scenarios, thereby adapting the pretrained natural domain representations to industrial domain-specific features. The introduction of these learnable weights represents a key departure from previous consistent self-attention approaches, which treat all heads equally and lack explicit domain adaptation capabilities.

The MHCSA block follows the standard transformer architecture with residual connections and layer normalization:

\begin{equation}
\begin{aligned}
\mathbf{Z}' &= \text{MHCSA}(\text{LN}(\mathbf{Z})) + \mathbf{Z}
\end{aligned}
\end{equation}

where $\mathbf{Z}$ denotes the input features, LN represents layer normalization.

\subsection{Text Encoder and Semantic Guidance}

To provide semantic guidance for anomaly detection, the text encoder from CLIP is employed to process textual descriptions of normal and abnormal states.

\textbf{Prompt Design.} Following the approach proposed in AnomalyCLIP \citep{anomalyclip}, learnable text prompts are employed to generate class-agnostic text representations that capture generic normality and abnormality patterns. Specifically, the prompt templates consist of learnable tokens combined with fixed object descriptors: $[\mathbf{V}]_1^M$ ``object'' for the normal state and $[\mathbf{W}]_1^M$ ``damaged object'' for the abnormal state, where $[\mathbf{V}]_1^M$ and $[\mathbf{W}]_1^M$ represent $M$ learnable token embeddings. Unlike conventional vision-language methods that use specific category names (e.g., ``bottle'', ``transistor''), these prompts employ the generic term ``object'' to create text embeddings that are invariant to object categories. The learnable tokens are optimized during training on auxiliary datasets to capture discriminative semantic features for distinguishing between normal and abnormal states. These learnable tokens are inserted into the first $L_t$ layers of the text encoder, enabling deep prompt tuning that refines the semantic representations throughout the text encoding process. The text encoder processes these prompts to generate semantic embeddings $\mathbf{F}_N \in \mathbb{R}^{D}$ for the normal state and $\mathbf{F}_A \in \mathbb{R}^{D}$ for the abnormal state, respectively.

\textbf{Vision-Language Alignment.} The text features guide the interpretation of visual features through cosine similarity computation. For each spatial location in the feature map, the similarity to both normal and abnormal text embeddings is computed, providing semantic context for anomaly scoring. The pretrained parameters of the text encoder remain frozen, while only the learnable token embeddings are optimized, preserving CLIP's semantic knowledge while adapting it for industrial anomaly detection.

\subsection{Joint Anomaly Score Computation}

As illustrated in Fig.~1(c), HeadCLIP proposes a JAS that combines global and local anomaly information to achieve robust detection performance. This represents a key contribution of the proposed framework, addressing limitations in existing vision-language-based anomaly detection methods that rely on either global-only or local-only information for image-level anomaly assessment.

\textbf{Global Anomaly Score.} The global anomaly score $S_g$ quantifies the image-level deviation from normality by measuring the similarity between the global visual representation and the abnormal text embedding. Following AnomalyCLIP~\citep{anomalyclip}, the score is computed as:

\begin{equation}
S_g = \text{sim}(\mathbf{F}_g, \mathbf{F}_A)
\end{equation}

where $\text{sim}(\cdot, \cdot)$ denotes cosine similarity, $\mathbf{F}_g$ represents the global image feature extracted from the CLS token of the global path, and $\mathbf{F}_A$ is the abnormal text embedding. The anomaly score increases when the global visual feature is more aligned with the abnormal text embedding, indicating the presence of anomalies.

\textbf{Local Anomaly Map.} The local anomaly map provides pixel-level anomaly localization by aggregating segmentation maps from multiple intermediate layers of the local path. Following the approach in AnomalyCLIP~\citep{anomalyclip}, normal and abnormal segmentation maps are computed at selected intermediate layers $\mathcal{M}$ and then merged. Specifically, for each layer $M_l \in \mathcal{M}$, the normal segmentation map $\mathbf{S}_{n,M_l}$ and abnormal segmentation map $\mathbf{S}_{a,M_l}$ are computed by measuring the similarity between local features and the normal/abnormal text embeddings, respectively. The final local anomaly map $\mathbf{S}_l \in \mathbb{R}^{H_{image} \times W_{image}}$ is obtained by:

\begin{equation}
\mathbf{S}_l = G_{\sigma}\left(\sum_{M_l \in \mathcal{M}} \left(\frac{1}{2}(\mathbf{I} - \text{Up}(\mathbf{S}_{n,M_l})) + \frac{1}{2}\text{Up}(\mathbf{S}_{a,M_l})\right)\right)
\end{equation}

where $\mathbf{I}$ is an identity matrix (all ones), $\text{Up}(\cdot)$ denotes bilinear upsampling to the original image resolution, and $G_{\sigma}$ represents a Gaussian filter with smoothing parameter $\sigma$. This formulation merges complementary information from multiple layers while applying spatial smoothing to reduce noise and produce coherent anomaly maps.

\textbf{Top-$k\%$ Sampling.} To mitigate the influence of noisy low-anomaly regions and extract the most discriminative local information, a top-$k\%$ sampling strategy is applied to the local anomaly map, selecting only the most anomalous pixels:

\begin{equation}
\bar{S}_l = \frac{1}{|\mathcal{K}|} \sum_{i \in \mathcal{K}} \mathbf{S}_l^{(i)}
\end{equation}

where $\mathcal{K}$ denotes the set of indices corresponding to the top $k\%$ highest anomaly scores, and $|\mathcal{K}|$ represents its cardinality. This aggregation strategy focuses on the most suspicious regions while filtering out noise from predominantly normal areas.

\textbf{Joint Anomaly Score.} The final JAS is computed by combining the global and local scores through a weighted sum:

\begin{equation}
S_j = r \cdot S_g + (1 - r) \cdot \bar{S}_l
\end{equation}

where $r \in [0, 1]$ is a balancing hyperparameter that controls the relative importance of global and local information. When $r$ is close to 1, the score emphasizes holistic semantic assessment, while values closer to 0 prioritize fine-grained spatial analysis. This weighted formulation synergistically leverages both semantic understanding from global features and fine-grained spatial analysis from local features. Unlike methods such as AnomalyCLIP~\citep{anomalyclip} and AdaCLIP~\citep{adaclip} that rely solely on global representations and may fail to detect small or localized anomalies, or methods like MSFlow~\citep{zhou2024msflow} that use only pixel-level information without holistic semantic context, the proposed JAS provides comprehensive anomaly assessment by integrating complementary information from both pathways. This design ensures robust detection across diverse anomaly types and scales, from large semantic deviations to subtle localized defects.

\subsection{Training Strategy with Auxiliary Datasets}

To enable effective learning of the learnable head weights and other trainable parameters, a training strategy employing auxiliary datasets with anomaly annotations is adopted.

\textbf{Auxiliary Dataset Selection.} Publicly available anomaly detection datasets (e.g., MVTec AD~\citep{mvtec}, VisA~\citep{visa}) are utilized as auxiliary data. These datasets provide diverse anomaly patterns across different object categories, enabling the model to learn generalizable anomaly representations. 

\textbf{Loss Functions.} The training objective consists of two components that supervise both global and local anomaly detection:

\textit{Global Loss} $\mathcal{L}_{\text{global}}$: This loss is a cross-entropy loss that matches the cosine similarity between the object-agnostic text embeddings and visual embeddings of normal/abnormal images from auxiliary data. It encourages the global anomaly score to distinguish between normal and abnormal samples at the image level:

\begin{equation}
\mathcal{L}_{\text{global}} = \mathbb{E}_{(\mathbf{I}, y)} \left[ \ell_{\text{CE}}(\text{sim}(\mathbf{F}_g, \mathbf{F}_N), \text{sim}(\mathbf{F}_g, \mathbf{F}_A), y) \right]
\end{equation}

where $y \in \{0, 1\}$ indicates normal or abnormal status, and $\ell_{\text{CE}}$ denotes the cross-entropy loss.

\textit{Local Loss} $\mathcal{L}_{\text{local}}^{M_l}$: For each selected intermediate layer $M_l \in \mathcal{M}$, the local loss supervises the pixel-level segmentation using ground truth masks $\mathbf{G} \in \{0, 1\}^{H_{image} \times W_{image}}$, where $\mathbf{G}_{jk} = 1$ if the pixel is anomalous and $\mathbf{G}_{jk} = 0$ otherwise. The local loss combines focal loss and Dice loss:

\begin{equation}
\begin{aligned}
\mathcal{L}_{\text{local}}^{M_l} = &\ell_{\text{Focal}}(\text{Up}([\mathbf{S}_{n,M_l}, \mathbf{S}_{a,M_l}]), \mathbf{G}) \\
&+ \ell_{\text{Dice}}(\text{Up}(\mathbf{S}_{n,M_l}), \mathbf{I} - \mathbf{G}) \\
&+ \ell_{\text{Dice}}(\text{Up}(\mathbf{S}_{a,M_l}), \mathbf{G})
\end{aligned}
\end{equation}

where $\ell_{\text{Focal}}(\cdot)$~\citep{lin2017focal} addresses class imbalance between normal and anomalous pixels by down-weighting easy examples, $\ell_{\text{Dice}}(\cdot)$~\citep{li2020dice} measures the overlap between predicted segmentation and ground truth mask to ensure accurate decision boundaries, $\text{Up}(\cdot)$ denotes upsampling operation, $[\cdot, \cdot]$ represents concatenation along the channel dimension, and $\mathbf{I}$ is an all-ones matrix. The focal loss is applied to the concatenated normal and abnormal segmentation maps, while the Dice losses separately measure the alignment of normal predictions with normal regions ($\mathbf{I} - \mathbf{S}$) and abnormal predictions with abnormal regions ($\mathbf{S}$).

The total training loss is formulated as:

\begin{equation}
\mathcal{L}_{\text{total}} = \mathcal{L}_{\text{global}} + \lambda \sum_{M_l \in \mathcal{M}} \mathcal{L}_{\text{local}}^{M_l}
\end{equation}

where $\lambda$ is a balancing hyperparameter that controls the relative importance between global and local losses, and the local losses are summed over all selected intermediate layers $\mathcal{M}$.

\textbf{Training Procedure.} During training, image-mask pairs are randomly sampled from auxiliary datasets, and only the learnable parameters (head weights and any additional lightweight modules) are optimized while keeping the CLIP backbone frozen. This approach ensures efficient training with limited computational resources while preserving the powerful semantic representations acquired from large-scale pretraining. After training on auxiliary data, the model can be directly applied to target industrial datasets without fine-tuning, demonstrating zero-shot generalization capabilities.

\section{Experiments}\label{Experiments}
\subsection{Experimental Settings}
The performance of HeadCLIP is evaluated on 17 publicly available anomaly detection datasets across two domains: industrial inspection and medical imaging.

\paragraph{Industrial domain} Seven datasets are employed that encompass both object-level and surface-level anomalies: MVTec AD \citep{mvtec}, VisA \citep{visa}, MPDD \citep{mpdd}, BTAD \citep{btad}, and SDD \citep{sdd} for object anomalies, and DAGM \citep{dagm} and DTD-Synthetic \citep{dtd} for surface anomalies.

\paragraph{Medical domain} Ten datasets are included covering a wide range of pathological conditions, including dermatology, endoscopy, radiology, and infectious diseases: ISIC \citep{isic}, CVC-ClinicDB \citep{clinicdb}, CVC-ColonDB \citep{colondb}, Kvasir \citep{kvasir}, Endo \citep{endo}, TN3k \citep{tn3k}, HeadCT \citep{headct&brainmri}, BrainMRI \citep{headct&brainmri}, Br35H \citep{br35h}, and COVID-19 \citep{covid19}. In the medical domain, different datasets are used for image-level and pixel-level evaluation due to the lack of datasets that simultaneously provide image-level labels and pixel-level anomaly masks. This is consistent with the settings adopted in previous methods such as AdaCLIP and AnomalyCLIP, and reflects the inherent limitations of publicly available medical datasets.

\paragraph{Evaluation metrics} Following the protocol in \citep{zhang2023exploring}, anomaly detection performance is assessed at two levels. For image-level evaluation, area under the receiver operating characteristic (ROC), average precision (AP), and maximum F1-score across thresholds ($\text{F1}_{\text{max}}$) are used. For pixel-level evaluation, AUROC, area under the per-region overlap curve (PRO) \citep{bergmann2020uninformed}, AP, F1-max, and maximum intersection over union ($\text{IoU}_{\text{max}}$) are employed, which provides order-independent evaluation of localization accuracy \citep{zhang2024learning}. Mean Anomaly Detection (mAD) \citep{zhang2023exploring}, computed as the average of all relevant metrics per level, is also reported to provide a comprehensive and balanced assessment. This multi-metric evaluation strategy ensures that both classification and localization capabilities are thoroughly examined under diverse operational conditions.

\begin{table}[!t]
\centering
\caption{Hyperparameter configuration for HeadCLIP training and inference.}
\label{tab:hyperparameters}
\small
\begin{adjustbox}{max width=\textwidth}
\begin{tabular}{lll}
\toprule
\textbf{Category} & \textbf{Hyperparameter} & \textbf{Value} \\
\midrule
\multicolumn{3}{l}{\textit{Training Configuration}} \\
\quad & Training split & Full training set (no validation split) \\
\quad & Epochs & 15 \\
\quad & Batch size & 8 \\
\quad & Random seed & 111 \\
\quad & Early stopping & Not used \\
\quad & Model selection & Final epoch (no target-domain signals) \\
\midrule
\multicolumn{3}{l}{\textit{Optimizer Settings}} \\
\quad & Optimizer & Adam \\
\quad & Learning rate & 0.001 \\
\quad & Betas ($\beta_1$, $\beta_2$) & (0.5, 0.999) \\
\quad & Weight decay & 0 \\
\quad & Learning rate scheduler & None \\
\midrule
\multicolumn{3}{l}{\textit{Prompt Learning}} \\
\quad & Visual prompt tokens ($E$) & 12 \\
\quad & Text prompt tokens & 4 \\
\quad & Prompt depth (text encoder layers) & 9 \\
\midrule
\multicolumn{3}{l}{\textit{Feature Extraction}} \\
\quad & Feature map layers & [0, 1, 2, 3] \\
\quad & ViT block indices & [6, 12, 18, 24] \\
\quad & CSA layer range & 5--19 \\
\quad & DPAM layers & 20 \\
\midrule
\multicolumn{3}{l}{\textit{Joint Anomaly Scoring (JAS)}} \\
\quad & Local score radius ($r$) & 3 \\
\quad & Top-$k$ ratio for global score & 0.1 \\
\quad & Temperature ($\tau$) & 0.07 (CLIP default) \\
\midrule
\multicolumn{3}{l}{\textit{Input Configuration}} \\
\quad & Image resolution & 518 $\times$ 518 \\
\quad & CLIP backbone & ViT-L/14@336px \\
\bottomrule
\end{tabular}
\end{adjustbox}
\end{table}

\subsection{Implementation Details}

HeadCLIP is built upon the publicly available CLIP model with ViT-L/14 architecture at 336px resolution. All backbone parameters remain frozen during training, with optimization restricted to three components: the learnable attention head weights in the image encoder, the learnable word embeddings appended to text prompts, and the learnable token embeddings inserted into early layers of the text encoder. The complete hyperparameter configuration is summarized in Table~\ref{tab:hyperparameters}.

For prompt learning, the number of learnable word embeddings $E$ is set to 12 for the visual prompt, while 4 learnable token embeddings are inserted at each of the first 9 transformer layers in the text encoder, following the protocol established by AnomalyCLIP~\citep{anomalyclip}. CSA is applied from layers 5 to 19 of the image encoder, with outputs from all CSA layers aggregated to form the local representation $F_{\text{local}}$. Each attention head within CSA is initialized with a learnable weight $w=1$ and optimized throughout training.

The model is trained for 15 epochs using the Adam optimizer with a learning rate of 0.001 and momentum coefficients $(\beta_1, \beta_2) = (0.5, 0.999)$. No learning rate scheduler or weight decay is employed. The batch size is set to 8, and input images are resized to $518 \times 518$ pixels. For Joint Anomaly Scoring, both the local score ratio $r$ and global score ratio $k$ are set to 0.5, with temperature $\tau$ fixed at the CLIP default value of 0.07.

Importantly, no early stopping or validation-based model selection is performed; the model checkpoint from the final training epoch is directly used for evaluation. This design ensures that no target-domain signals influence the training process, preserving the integrity of the zero-shot evaluation protocol.

Two cross-domain experimental protocols are adopted to evaluate generalization capability. In the industrial pretraining protocol, the model is trained on VisA and evaluated on MVTec AD. In the generalization testing protocol, the model is trained on MVTec AD and evaluated on all remaining datasets across both industrial and medical domains.

HeadCLIP is benchmarked against existing ZSAD approaches, including AnoVL~\citep{anovl}, AnomalyCLIP~\citep{anomalyclip}, and AdaCLIP~\citep{adaclip}. All baseline results are reproduced using official codebases with default settings to ensure fair comparison. Experiments are conducted using PyTorch 2.2.0 on a single NVIDIA RTX 2080 GPU with 12GB memory. For datasets containing multiple sub-categories, reported scores represent the average performance across all sub-categories.

\subsection{Zero-shot Anomaly Detection Results}

\begin{table}[!ht]
\caption{ZSAD performance comparison in industrial domain at image and pixel levels. Best performance shown in bold, second-best underlined.}
\label{tab:main_industrial}
\centering
\begin{adjustbox}{max width=\textwidth}
\begin{tabular}{@{}cl*{10}{w{c}{1.2cm}}@{}}
\toprule
\multirow{2}{*}{Dataset} &
  \multicolumn{1}{l}{\multirow{2}{*}{Model}} &
  \multicolumn{4}{c}{Image-level} &
  \multicolumn{6}{c}{Pixel-level} \\ \cmidrule(lr){3-6} \cmidrule(lr){7-12}
 &
  \multicolumn{1}{c}{} &
  ROC &
  AP &
  $\text{F1}_{\text{max}}$ &
  mAD &
  ROC &
  PRO &
  AP &
  $\text{F1}_{\text{max}}$ &
  $\text{IoU}_{\text{max}}$ &
  mAD \\ \midrule
\multirow{4}{*}{\rotatebox{90}{MVTecAD}} &
  AnoVL &
  91.5 &
  96.4 &
  92.8 &
  93.6 &
  88.6 &
  74.6 &
  27.2 &
  32.9 &
  20.4 &
  48.7 \\
 &
  AnomalyCLIP &
  90.2 &
  95.7 &
  92.2 &
  92.7 &
  89.2 &
  {\ul 81.9} &
  34.7 &
  {\ul 37.8} &
  {\ul 24.6} &
  {\ul 53.6} \\
 &
  AdaCLIP &
  {\ul 92.9} &
  \textbf{96.9} &
  {\ul 92.9} &
  {\ul 94.2} &
  {\ul 90.3} &
  36.2 &
  \textbf{43.2} &
  24.7 &
  14.9 &
  41.9 \\
 &
  HeadCLIP &
  \textbf{93.2} &
  {\ul 96.8} &
  \textbf{93.4} &
  \textbf{94.5} &
  \textbf{92.0} &
  \textbf{82.2} &
  {\ul 39.9} &
  \textbf{40.6} &
  \textbf{26.7} &
  \textbf{56.3} \\ \midrule
\multirow{4}{*}{\rotatebox{90}{VisA}} &
  AnoVL &
  76.8 &
  79.4 &
  78.8 &
  78.3 &
  88.2 &
  66.4 &
  10.4 &
  15.4 &
  9.1 &
  37.9 \\
 &
  AnomalyCLIP &
  80.9 &
  83.9 &
  79.6 &
  81.5 &
  94.0 &
  {\ul 86.5} &
  20.2 &
  25.7 &
  16.1 &
  {\ul 48.5} \\
 &
  AdaCLIP &
  {\ul 85.4} &
  \textbf{88.3} &
  {\ul 83.2} &
  {\ul 85.6} &
  {\ul 95.6} &
  48.3 &
  \textbf{29.1} &
  \textbf{31.0} &
  \textbf{19.9} &
  44.8 \\
 &
  HeadCLIP &
  \textbf{86.3} &
  {\ul 88.2} &
  \textbf{83.4} &
  \textbf{86.0} &
  \textbf{95.8} &
  \textbf{88.8} &
  {\ul 23.4} &
  {\ul 30.6} &
  {\ul 19.1} &
  \textbf{51.5} \\ \midrule
\multirow{4}{*}{\rotatebox{90}{MPDD}} &
  AnoVL &
  65.0 &
  71.9 &
  78.3 &
  71.8 &
  65.0 &
  36.4 &
  13.9 &
  15.0 &
  10.6 &
  28.2 \\
 &
  AnomalyCLIP &
  72.8 &
  {\ul 79.8} &
  79.3 &
  77.3 &
  {\ul 96.1} &
  \textbf{89.8} &
  23.5 &
  28.4 &
  18.0 &
  {\ul 51.2} \\
 &
  AdaCLIP &
  {\ul 76.6} &
  \textbf{80.5} &
  {\ul 81.0} &
  \textbf{79.4} &
  \textbf{96.4} &
  36.7 &
  \textbf{31.1} &
  {\ul 28.6} &
  \textbf{18.8} &
  42.3 \\
 &
  HeadCLIP &
  \textbf{77.4} &
  79.4 &
  \textbf{81.1} &
  {\ul 79.3} &
  95.7 &
  {\ul 87.9} &
  {\ul 25.9} &
  \textbf{29.7} &
  {\ul 18.7} &
  \textbf{51.6} \\ \midrule
\multirow{4}{*}{\rotatebox{90}{BTAD}} &
  AnoVL &
  72.0 &
  68.2 &
  69.3 &
  69.8 &
  86.6 &
  51.2 &
  23.4 &
  29.2 &
  18.1 &
  41.7 \\
 &
  AnomalyCLIP &
  {\ul 87.1} &
  82.3 &
  79.2 &
  82.8 &
  90.9 &
  {\ul 67.9} &
  31.2 &
  {\ul 34.0} &
  {\ul 22.5} &
  {\ul 49.3} \\
 &
  AdaCLIP &
  78.4 &
  {\ul 89.9} &
  {\ul 87.9} &
  {\ul 85.4} &
  {\ul 92.0} &
  25.8 &
  {\ul 46.4} &
  27.3 &
  16.0 &
  41.5 \\
 &
  HeadCLIP &
  \textbf{94.5} &
  \textbf{97.2} &
  \textbf{93.9} &
  \textbf{95.2} &
  \textbf{95.6} &
  \textbf{81.4} &
  \textbf{47.7} &
  \textbf{50.9} &
  \textbf{35.6} &
  \textbf{62.2} \\ \midrule
\multirow{4}{*}{\rotatebox{90}{SDD}} &
  AnoVL &
  83.2 &
  76.9 &
  67.2 &
  75.8 &
  {\ul 86.7} &
  {\ul 53.6} &
  19.2 &
  27.7 &
  16.1 &
  40.7 \\
 &
  AnomalyCLIP &
  \textbf{85.4} &
  \textbf{80.8} &
  {\ul 74.0} &
  \textbf{80.0} &
  82.1 &
  53.0 &
  {\ul 28.1} &
  {\ul 35.5} &
  {\ul 21.6} &
  {\ul 44.1} \\
 &
  AdaCLIP &
  75.8 &
  60.7 &
  57.1 &
  64.5 &
  73.9 &
  5.9 &
  21.4 &
  7.0 &
  3.6 &
  26.2 \\
 &
  HeadCLIP &
  {\ul 84.6} &
  {\ul 77.2} &
  \textbf{74.2} &
  {\ul 78.7} &
  \textbf{91.3} &
  \textbf{65.6} &
  \textbf{40.7} &
  \textbf{43.3} &
  \textbf{27.6} &
  \textbf{53.7} \\ \midrule
\multirow{4}{*}{\rotatebox{90}{DAGM}} &
  AnoVL &
  89.9 &
  75.1 &
  72.6 &
  79.2 &
  87.5 &
  69.0 &
  11.9 &
  17.8 &
  10.4 &
  39.3 \\
 &
  AnomalyCLIP &
  97.4 &
  91.7 &
  89.7 &
  92.9 &
  93.6 &
  {\ul 87.2} &
  53.1 &
  {\ul 55.4} &
  {\ul 40.1} &
  {\ul 65.9} \\
 &
  AdaCLIP &
  {\ul 98.6} &
  {\ul 94.8} &
  {\ul 92.6} &
  {\ul 95.3} &
  {\ul 95.3} &
  28.6 &
  \textbf{61.6} &
  27.9 &
  17.3 &
  46.1 \\
 &
  HeadCLIP &
  \textbf{99.2} &
  \textbf{96.9} &
  \textbf{94.5} &
  \textbf{96.9} &
  \textbf{96.2} &
  \textbf{92.4} &
  61.0 &
  \textbf{61.0} &
  \textbf{45.7} &
  \textbf{71.3} \\ \midrule
\multirow{4}{*}{\rotatebox{90}{\shortstack{DTD\\-Synthetic}}} &
  AnoVL &
  93.7 &
  96.9 &
  92.7 &
  94.4 &
  96.4 &
  87.0 &
  37.0 &
  40.9 &
  26.3 &
  57.5 \\
 &
  AnomalyCLIP &
  93.0 &
  96.9 &
  93.6 &
  94.5 &
  98.3 &
  \textbf{94.3} &
  71.8 &
  \textbf{67.8} &
  \textbf{51.8} &
  {\ul 76.8} \\
 &
  AdaCLIP &
  {\ul 96.4} &
  {\ul 98.3} &
  {\ul 95.6} &
  {\ul 96.8} &
  {\ul 98.3} &
  58.1 &
  \textbf{76.6} &
  {\ul 59.4} &
  43.9 &
  67.2 \\
 &
  HeadCLIP &
  \textbf{96.9} &
  \textbf{98.8} &
  \textbf{95.7} &
  \textbf{97.1} &
  \textbf{98.6} &
  {\ul 94.0} &
  {\ul 72.6} &
  \textbf{67.8} &
  {\ul 51.7} &
  \textbf{76.9} \\ \midrule
\multirow{4}{*}{\rotatebox{90}{Average}} &
  AnoVL &
  81.7 &
  80.7 &
  78.8 &
  80.4 &
  85.6 &
  62.6 &
  20.4 &
  25.6 &
  15.8 &
  42.0 \\
 &
  AnomalyCLIP &
  {\ul 86.7} &
  {\ul 87.3} &
  84.0 &
  {\ul 86.0} &
  {\ul 92.0} &
  {\ul 80.1} &
  37.5 &
  {\ul 40.7} &
  {\ul 27.8} &
  {\ul 55.6} \\
 &
  AdaCLIP &
  86.3 &
  87.1 &
  {\ul 84.3} &
  85.9 &
  91.7 &
  34.2 &
  {\ul 44.2} &
  29.4 &
  19.2 &
  44.3 \\
 &
  HeadCLIP &
  \textbf{90.3} &
  \textbf{90.6} &
  \textbf{88.0} &
  \textbf{89.7} &
  \textbf{95.0} &
  \textbf{84.6} &
  \textbf{44.5} &
  \textbf{46.3} &
  \textbf{32.2} &
  \textbf{60.5} \\ \bottomrule
\end{tabular}
\end{adjustbox}
\end{table}

\begin{table}[!ht]
\caption{ZSAD performance comparison in medical domain at pixel level. Best performance shown in bold, second-best underlined.}
\label{tab:main_medical_pixel}
\centering
\small
\begin{adjustbox}{max width=\textwidth}
\begin{tabular}{@{}cl*{6}{w{c}{1.2cm}}@{}}
\toprule
\multirow{2}{*}{Dataset}      & \multicolumn{1}{c}{\multirow{2}{*}{Model}} & \multicolumn{6}{c}{Pixel-level}                                                               \\ \cmidrule(l){3-8} 
                              & \multicolumn{1}{c}{}                       & ROC           & PRO           & AP            & $\text{F1}_{\text{max}}$       & $\text{IoU}_{\text{max}}$      & mAD           \\ \midrule
\multirow{4}{*}{ISIC} & AnoVL    & \textbf{93.3}       & \textbf{84.2} & \textbf{84.8} & \textbf{77.4} & \textbf{63.1} & \textbf{80.6} \\
                              & AnomalyCLIP                                & {\ul 90.6}    & 81.5          & {\ul 79.1}    & {\ul 72.9}    & {\ul 57.3}    & {\ul 76.3}    \\
                              & AdaCLIP                                    & 88.3          & 31.0          & 76.5          & 44.0          & 28.2          & 53.6          \\
                              & HeadCLIP                                   & 90.1          & {\ul 81.7}    & 78.8          & 72.1          & 56.4          & 75.8          \\ \midrule
\multirow{4}{*}{CVC-ColonDB}  & AnoVL                                      & 77.9          & 48.9          & 21.2          & 28.9          & 16.9          & 38.8          \\
                              & AnomalyCLIP                                & {\ul 81.6}    & {\ul 69.1}    & {\ul 25.6}    & {\ul 34.1}    & {\ul 20.6}    & {\ul 46.2}    \\
                              & AdaCLIP                                    & 80.8          & 44.9          & 24.3          & 27.8          & 16.1          & 38.8          \\
                              & HeadCLIP                                   & \textbf{83.8} & \textbf{72.8} & \textbf{30.3} & \textbf{37.1} & \textbf{22.8} & \textbf{49.4} \\ \midrule
\multirow{4}{*}{CVC-ClinicDB} & AnoVL                                      & 80.9          & 53.0          & {\ul 32.7}    & 37.0          & 22.7          & 45.3          \\
                              & AnomalyCLIP                                & {\ul 82.5}    & {\ul 65.8}    & 31.6          & {\ul 39.3}    & {\ul 24.4}    & {\ul 48.7}    \\
                              & AdaCLIP                                    & 77.0          & 26.5          & 27.3          & 25.9          & 14.9          & 34.3          \\
                      & HeadCLIP & {\ul \textbf{85.6}} & \textbf{71.0} & \textbf{38.1} & \textbf{44.2} & \textbf{28.4} & \textbf{53.5} \\ \midrule
\multirow{4}{*}{Kvasir}       & AnoVL                                      & 73.4          & 26.2          & 32.4          & 38.8          & 24.1          & 39.0          \\
                              & AnomalyCLIP                                & {\ul 79.0}    & \textbf{50.5} & {\ul 38.4}    & {\ul 45.4}    & {\ul 29.4}    & {\ul 48.5}    \\
                              & AdaCLIP                                    & 76.4          & 40.1          & 32.5          & 33.9          & 20.4          & 40.7          \\
                              & HeadCLIP                                   & \textbf{82.8} & {\ul 48.0}    & \textbf{43.2} & \textbf{50.0} & \textbf{33.3} & \textbf{51.5} \\ \midrule
\multirow{4}{*}{Endo}         & AnoVL                                      & 81.1          & 47.5          & 38.9          & 44.7          & 28.8          & 48.2          \\
                              & AnomalyCLIP                                & {\ul 82.9}    & {\ul 60.2}    & {\ul 42.6}    & {\ul 47.6}    & {\ul 31.2}    & {\ul 52.9}    \\
                              & AdaCLIP                                    & 81.6          & 44.2          & 38.2          & 38.1          & 23.5          & 45.1          \\
                              & HeadCLIP                                   & \textbf{86.1} & \textbf{66.8} & \textbf{48.1} & \textbf{52.2} & \textbf{35.3} & \textbf{57.7} \\ \midrule
\multirow{4}{*}{TN3K}         & AnoVL                                      & 74.9          & 39.4          & 29.0          & 35.9          & 21.9          & 40.2          \\
                              & AnomalyCLIP                                & 76.8          & {\ul 45.2}    & 38.7          & {\ul 40.5}    & {\ul 25.4}    & {\ul 45.3}    \\
                              & AdaCLIP                                    & {\ul 80.6}    & 17.0          & {\ul 42.9}    & 29.2          & 17.1          & 37.4          \\
                              & HeadCLIP                                   & \textbf{80.9} & \textbf{46.4} & \textbf{44.3} & \textbf{45.9} & \textbf{29.8} & \textbf{49.5} \\ \midrule
\multirow{4}{*}{Average}      & AnoVL                                      & 80.3          & 49.9          & 39.8          & 43.8          & 29.6          & 48.7          \\
                              & AnomalyCLIP                                & {\ul 82.2}    & {\ul 62.1}    & {\ul 42.7}    & {\ul 46.6}    & {\ul 31.4}    & {\ul 53.0}    \\
                              & AdaCLIP                                    & 80.8          & 34.0          & 40.3          & 33.1          & 20.0          & 41.6          \\
                              & HeadCLIP                                   & \textbf{84.9} & \textbf{64.4} & \textbf{47.2} & \textbf{50.2} & \textbf{34.3} & \textbf{56.2} \\ \bottomrule
\end{tabular}
\end{adjustbox}
\end{table}

\begin{table}[!ht]
\caption{ZSAD performance comparison in medical domain at image level. Best performance shown in bold, second-best underlined}
\label{tab:main_medical_image}
\centering
\small
\begin{adjustbox}{max width=\textwidth}
\begin{tabular}{@{}cl*{4}{w{c}{1.2cm}}@{}}
\toprule
\multirow{2}{*}{Dataset}  & \multirow{2}{*}{Model} & \multicolumn{4}{c}{Image-level}                               \\ \cmidrule(l){3-6} 
                          &                        & ROC           & AP            & $\text{F1}_{\text{max}}$       & mAD           \\ \midrule
\multirow{4}{*}{HeadCT}   & AnoVL                  & 85.1          & 84.8          & 79.0          & 83.0          \\
                          & AnomalyCLIP            & 91.8          & 91.5          & 87.9          & 90.4          \\
                          & AdaCLIP                & {\ul 94.7}    & {\ul 93.7}    & {\ul 89.4}    & {\ul 92.6}    \\
                          & HeadCLIP               & \textbf{95.3} & \textbf{95.9} & \textbf{91.5} & \textbf{94.2} \\ \midrule
\multirow{4}{*}{BrainMRI} & AnoVL                  & 85.2          & 90.4          & 83.8          & 86.5          \\
                          & AnomalyCLIP            & 92.3          & 94.1          & 89.9          & 92.1          \\
                          & AdaCLIP                & \textbf{96.5} & \textbf{97.6} & \textbf{94.3} & \textbf{96.1} \\
                          & HeadCLIP               & {\ul 96.1}    & {\ul 96.5}    & {\ul 93.8}    & {\ul 95.5}    \\ \midrule
\multirow{4}{*}{Br35H}    & AnoVL                  & 82.7          & 83.4          & 77.4          & 81.2          \\
                          & AnomalyCLIP            & 94.1          & 94.5          & 86.9          & 91.8          \\
                          & AdaCLIP                & \textbf{98.0} & \textbf{98.1} & {\ul 92.2}    & {\ul 96.1}    \\
                          & HeadCLIP               & {\ul 97.9}    & {\ul 97.9}    & \textbf{93.3} & \textbf{96.4} \\ \midrule
\multirow{4}{*}{COVID-19} & AnoVL                  & 64.2          & 44.1          & 42.4          & 50.2          \\
                          & AnomalyCLIP            & {\ul 79.3}    & {\ul 53.4}    & {\ul 53.6}    & {\ul 62.1}    \\
                          & AdaCLIP                & 70.8          & 48.4          & 49.0          & 56.1          \\
                          & HeadCLIP               & \textbf{86.5} & \textbf{59.2} & \textbf{56.4} & \textbf{67.3} \\ \midrule
\multirow{4}{*}{Average}  & AnoVL                  & 79.3          & 75.7          & 70.6          & 75.2          \\
                          & AnomalyCLIP            & 89.4          & 83.4          & 79.6          & 84.1          \\
                          & AdaCLIP                & {\ul 90.0}    & {\ul 84.5}    & {\ul 81.2}    & {\ul 85.2}    \\
                          & HeadCLIP               & \textbf{93.9} & \textbf{87.4} & \textbf{83.7} & \textbf{88.4} \\ \bottomrule
\end{tabular}
\end{adjustbox}
\end{table}

The performance of HeadCLIP in ZSAD is evaluated across both industrial and medical domains. The results demonstrate that HeadCLIP consistently outperforms prior state-of-the-art methods in both pixel-level and image-level detection tasks.
\paragraph{Industrial domain}
As summarized in Table~\ref{tab:main_industrial}, HeadCLIP achieves a 4.9\%p increase in mAD for pixel-level detection compared to the previous best-performing methods. The model attains the highest mAD across all seven industrial datasets and achieves the highest average score across six evaluation metrics. When considering 35 evaluation cases across 7 datasets and 5 metrics, HeadCLIP ranks first in 22 cases and second in 12 cases. For image-level detection, HeadCLIP improves mAD by 3.7\%p, ranking first in 5 out of 7 datasets and showing the highest average across all four metrics. When examining 21 image-level evaluation cases across 7 datasets and 3 metrics, HeadCLIP ranks first in 16 cases and second in 6 cases.

\paragraph{Medical domain}
Table~\ref{tab:main_medical_pixel} presents the pixel-level performance on six medical datasets. HeadCLIP improves overall mAD by 3.2\%p, achieving the highest average performance across all six metrics. When considering 30 evaluation cases across 6 datasets and 5 metrics, the model ranks first in 24 cases and second in 2 cases. In image-level detection shown in Table~\ref{tab:main_medical_image}, HeadCLIP similarly achieves a 3.2\%p improvement in mAD, ranking first in 7 out of 12 evaluation cases across 4 datasets and 3 metrics, and second in 5 cases.

\begin{table}[!ht]
\caption{Comparison with recent CLIP-based zero-shot anomaly detection methods. Performance is evaluated across industrial and medical domains using both image-level metrics and pixel-level metrics. \textbf{Bold} indicates the best performance and \underline{underline} indicates the second-best performance for each dataset.}
\label{tab:main_sota}
\centering
\begin{adjustbox}{max width=\textwidth}
\begin{tabular}{@{}cccccc@{}}
\toprule
Domain & Metric                                                                                & Dataset       & AA-CLIP      & AF-CLIP      & HeadCLIP     \\ \midrule
\multirow{10}{*}{Industrial} & \multirow{5}{*}{\begin{tabular}[c]{@{}c@{}}Image-level\\ (ROC, AP)\end{tabular}} & MVTecAD  & (90.5, 94.9) & (92.9, 96.8) & (\textbf{93.2}, \textbf{96.8}) \\
       &                                                                                       & VisA          & (84.6, 82.2) & (\textbf{88.5}, \textbf{90.0}) & ({\ul 86.3}, {\ul 88.2}) \\
       &                                                                                       & BTAD          & (93.8, \textbf{97.9}) & ({\ul 94.3}, 95.2) & (\textbf{94.5}, {\ul 97.2}) \\
       &                                                                                       & DAGM          & (93.9, 84.5) & ({\ul 98.7}, {\ul 94.6}) & (\textbf{99.2}, \textbf{96.9}) \\
       &                                                                                       & DTD-Synthetic & (93.3, 97.8) & (\textbf{97.9}, \textbf{99.1}) & ({\ul 96.9}, {\ul 98.8}) \\ \cmidrule(l){2-6} 
       & \multirow{5}{*}{\begin{tabular}[c]{@{}c@{}}Pixel-level\\ (ROC, PRO)\end{tabular}} & MVTec         & (91.9, {\ul 84.6}) & (\textbf{92.3}, \textbf{85.7}) & ({\ul 92.0}, 82.2) \\
       &                                                                                       & VisA          & (95.5, 83.0) & (\textbf{96.2}, {\ul 88.7}) & ({\ul 95.8}, \textbf{88.8}) \\
       &                                                                                       & BTAD          & (94.0, 69.0) & ({\ul 94.4}, {\ul 78.3}) & (\textbf{95.6}, \textbf{81.4}) \\
       &                                                                                       & DAGM          & (91.6, 76.5) & (\textbf{97.1}, \textbf{93.1}) & ({\ul 96.2}, {\ul 92.4}) \\
       &                                                                                       & DTD-Synthetic & (96.4, 85.9) & ({\ul 98.6}, {\ul 93.8}) & (\textbf{98.6}, \textbf{94.0}) \\ \midrule
\multirow{6}{*}{Medical}     & \multirow{2}{*}{\begin{tabular}[c]{@{}c@{}}Image-level\\ (ROC, AP)\end{tabular}} & BrainMRI & (91.8, 94.1) & ({\ul 95.2}, {\ul 96.3}) & (\textbf{96.1}, \textbf{96.5}) \\
       &                                                                                       & Br35H         & (89.4, 91.0) & ({\ul 96.7}, {\ul 96.4}) & (\textbf{97.9}, \textbf{97.9}) \\ \cmidrule(l){2-6} 
       & \multirow{4}{*}{\begin{tabular}[c]{@{}c@{}}Pixel-level\\ (ROC, PRO)\end{tabular}} & ISIC          & ({\ul 93.9}, {\ul 87.0}) & (\textbf{94.8}, \textbf{89.6})  & (90.1, 81.7) \\
       &                                                                                       & CVC-ColonDB   & (81.6, 65.4) & ({\ul 83.2}, {\ul 67.9}) & (\textbf{83.8}, \textbf{72.8}) \\
       &                                                                                       & CVC-ClinicDB  & ({\ul 86.3}, 67.1) & (\textbf{87.1}, {\ul 70.0}) & (85.6, \textbf{71.0}) \\
       &                                                                                       & Kvasir        & ({\ul 84.0}, 50.3) & (\textbf{84.5}, \textbf{57.5}) & (82.8, 48.0) \\ \bottomrule
\end{tabular}
\end{adjustbox}
\end{table}

\paragraph{Comparison with Recent CLIP-based ZSAD Methods}
Table~\ref{tab:main_sota} compares HeadCLIP with recent state-of-the-art CLIP-based zero-shot anomaly detection methods, AA-CLIP and AF-CLIP, across both industrial and medical domains. HeadCLIP achieves competitive or superior performance across both domains, demonstrating the effectiveness of learnable head weights for domain adaptation. In the industrial domain, HeadCLIP achieves the best performance on multiple datasets including DAGM with 99.2\% ROC-AUC for image-level detection and 96.2\% for pixel-level detection, and matches state-of-the-art results on others such as DTD-Synthetic with 98.6\% ROC-AUC for pixel-level detection. In the medical domain, HeadCLIP demonstrates strong cross-domain generalization despite being trained exclusively on industrial data, achieving the best performance on BrainMRI with 96.1\% ROC-AUC and Br35H with 97.9\% ROC-AUC for image-level detection. These results validate that HeadCLIP's learnable head weights provide an effective and parameter-efficient approach for adapting CLIP to specialized anomaly detection tasks, achieving performance comparable to or exceeding recent ZSAD methods across diverse visual domains.

\begin{figure}[!ht]
    \centering
    \includegraphics[width=1\linewidth]{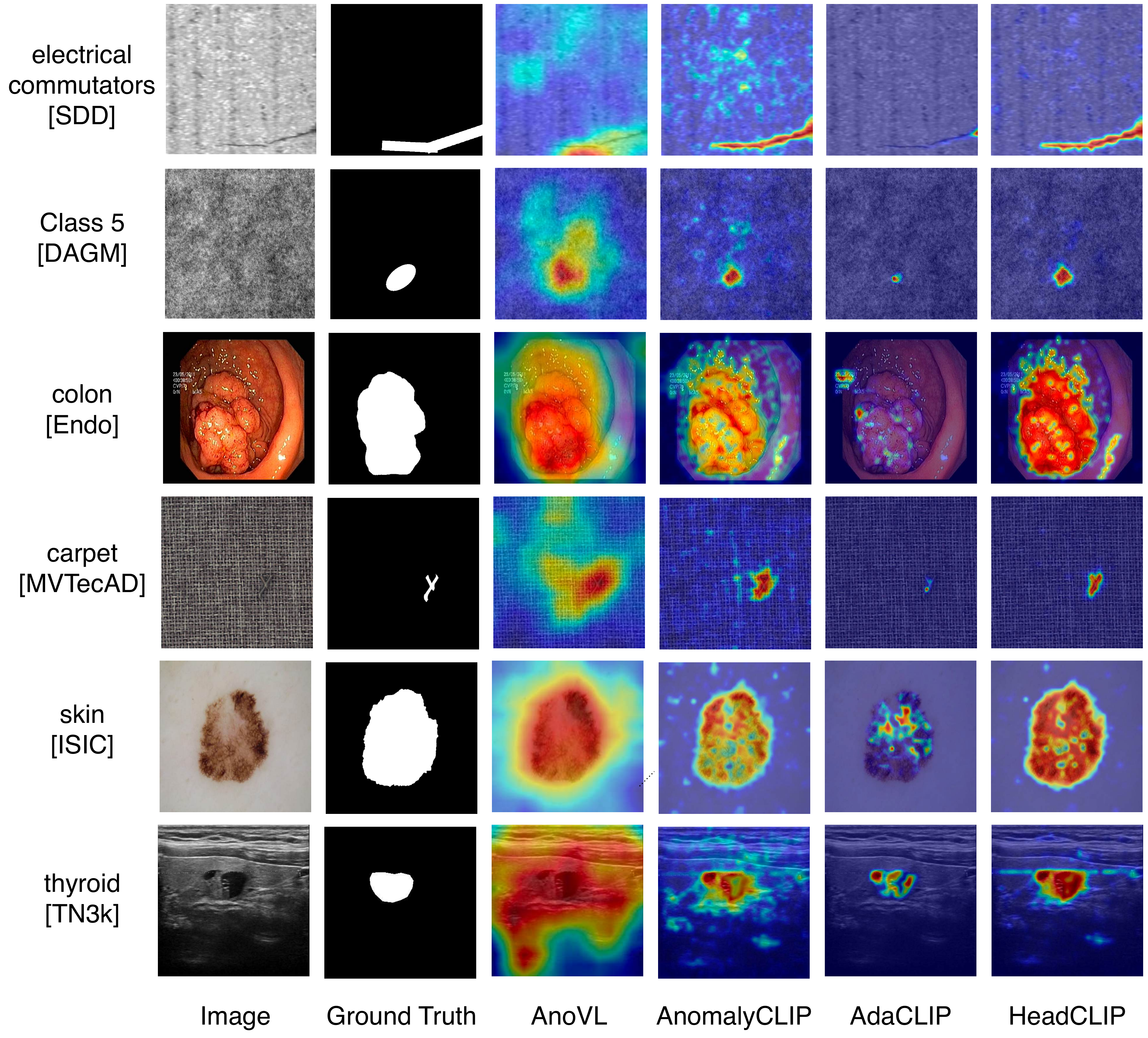}
    \vspace{-20pt}
    \caption{Qualitative comparison of pixel-level anomaly detection results across different models. The leftmost column shows original input images from various domains, along with their corresponding ground truth masks. The comparison demonstrates that HeadCLIP achieves more precise anomaly localization and generates clearer anomaly maps compared to previous methods. These results validate the effectiveness of the proposed method in achieving robust anomaly detection across diverse domains.}
    \label{fig:main_qualitative}
\end{figure}

\paragraph{Qualitative comparison}

Figure~\ref{fig:main_qualitative} illustrates anomaly localization results. HeadCLIP produces cleaner, more localized anomaly maps with reduced false positives compared to prior approaches, in both industrial and medical settings.

\paragraph{Analysis}
The superior performance of HeadCLIP can be attributed to two primary design choices: (1) the LHW enable fine-grained, head-level adaptation of the image encoder to the target domain, resulting in more accurate anomaly localization, and (2) the JAS effectively fuses pixel-level and image-level anomaly evidence, leveraging domain-adapted features from the local path to improve global detection robustness. Due to space limitations, detailed class-wise performance results for each dataset are provided in Appendix. Together, these components contribute to HeadCLIP's stable and generalizable performance across varied datasets and domains.

\begin{figure}[!ht]
    \centering
    \includegraphics[width=1\linewidth]{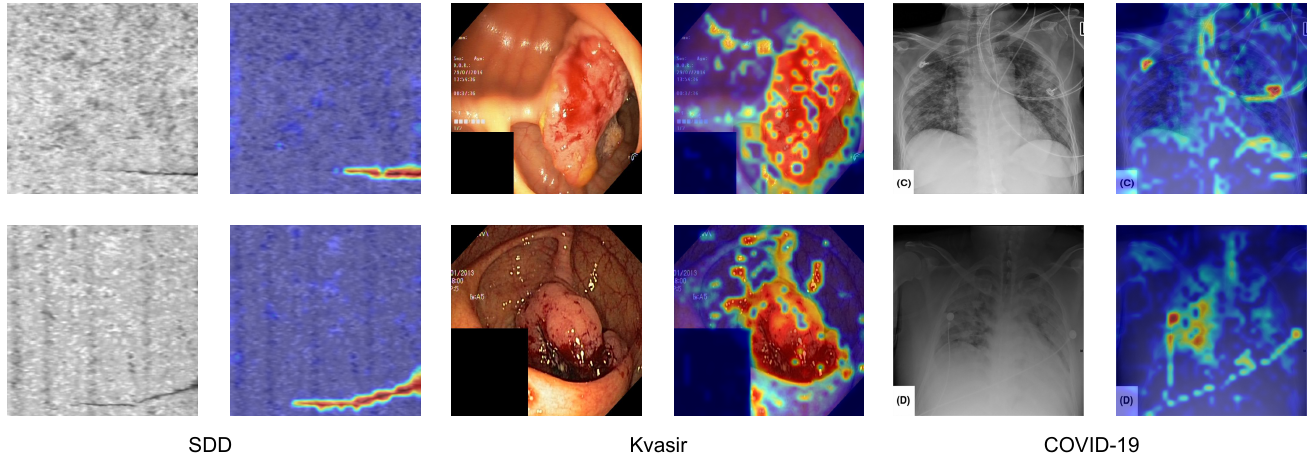}
    \vspace{-30pt}
    \caption{Qualitative examples demonstrating HeadCLIP's robustness to real-world challenges. The left column (SDD) shows that HeadCLIP successfully detects subtle surface defects under low contrast, low resolution, and high irregularity conditions that introduce significant visual noise. The middle column (Kvasir) illustrates that black rectangular occlusions near the image boundary do not interfere with detection, as these patterns are semantically familiar from training data. The right column (COVID-19) highlights that semantically unusual features, such as metallic lines wrapped around the patient's body, are identified as anomalies, reflecting the semantic prior-based nature of zero-shot detection.}
    \label{fig:realworld_qualitative}
\end{figure}

\paragraph{Robustness to Real-World Challenges}
To evaluate HeadCLIP's practicality in real-world environments, its robustness under common challenges such as distribution shift, noise, and occlusion is examined. As a ZSAD framework grounded in semantic understanding, HeadCLIP detects anomalies based on learned concepts of normality and abnormality, rather than relying solely on low-level visual differences.

\begin{itemize}
    \item \textbf{Distribution shift.} HeadCLIP is trained exclusively on MVTec AD and evaluated on 16 diverse datasets spanning industrial and medical domains, covering various modalities and anomaly types. Despite the significant distribution gap, it consistently achieves state-of-the-art performance in both pixel-level and image-level tasks (Tables~\ref{tab:main_industrial} and~\ref{tab:main_medical_pixel}), demonstrating strong generalization to unseen categories.

    \item \textbf{Noise robustness.} The SDD dataset exhibits significant visual noise in the form of low contrast, low resolution, and high level of background irregularity. These characteristics make it difficult to distinguish anomalies such as surface cracks from the background. Despite these challenges, HeadCLIP produces clean and semantically coherent anomaly maps (Figure~\ref{fig:realworld_qualitative} - SDD), successfully detecting subtle defects while suppressing irrelevant background variation. This demonstrates the model's robustness to noisy conditions, highlighting its ability to rely on learned semantic cues rather than low-level texture fluctuations.

    \item \textbf{Occlusion robustness.} It is observed that HeadCLIP's anomaly decisions are influenced not by the mere presence of occlusion, but by the semantic familiarity of the occluded content. For example, in the Kvasir dataset, black rectangular patches near the lower left corner do not trigger false detections because dark peripheral regions are common in the MVTec AD training set. (Figure~\ref{fig:realworld_qualitative} - Kvasir) In contrast, in the COVID-19 dataset, metallic lines wrapped around a patient's body are frequently detected as anomalies, since such visual patterns are semantically unusual and unfamiliar. (Figure~\ref{fig:realworld_qualitative} - COVID-19) This behavior highlights a defining characteristic of ZSAD models: anomaly decisions are driven by prior knowledge of what constitutes semantically normal or abnormal, rather than low-level deviations alone.
\end{itemize}

These findings demonstrate HeadCLIP's robustness to various real-world conditions while also revealing the inherent trade-offs of semantic prior-based detection. Future work may explore techniques to incorporate context-awareness or adaptive priors to further mitigate sensitivity to benign yet uncommon patterns.

\subsection{Learnable Head Weights Analysis}
To better understand the operational characteristics and generalization behavior of the proposed LHW, a detailed analysis is conducted along four dimensions:
\begin{itemize}
    \item Effectiveness in domain adaptation: Whether LHW facilitates adaptation of the image encoder to target domains is examined, and how this contributes to anomaly detection performance.
    \item Robustness across feature extraction layers: Whether LHW yields consistent improvements across various layer combinations used to construct $F_{\text{local}}$ is assessed.
    \item Sensitivity to initialization (described in the following section): How different initial values for LHW impact convergence and final performance is analyzed.
    
    \item Interpretability analysis: The distribution of learned head weights is visualized to analyze generalizable and dataset-specific attention patterns.
    
\end{itemize}

These analyses provide insight into the underlying mechanisms of LHW and inform its optimal configuration.

\begin{table}[!ht]
\caption{Ablation study results of LHW in industrial domain at image and pixel levels without JAS. Higher performance shown in bold.}
\label{tab:LHW_ablation_industrial}
\centering
\begin{adjustbox}{max width=\textwidth}
\begin{tabular}{@{}cc*{10}{w{c}{1.2cm}}@{}}
\toprule
\multirow{2}{*}{Dataset} &
  \multirow{2}{*}{LHW} &
  \multicolumn{4}{c}{Image-level} &
  \multicolumn{6}{c}{Pixel-level} \\ \cmidrule(lr){3-6} \cmidrule(lr){7-12}
 &
  &
  ROC &
  AP &
  $\text{F1}_{\text{max}}$ &
  mAD &
  ROC &
  PRO &
  AP &
  $\text{F1}_{\text{max}}$ &
  $\text{IoU}_{\text{max}}$ &
  mAD \\ \midrule
\multirow{2}{*}{MVTecAD} &
  &
  \textbf{90.7} &
  95.6 &
  \textbf{92.4} &
  \textbf{92.9} &
  90.5 &
  81.8 &
  32.4 &
  36.0 &
  23.4 &
  52.8 \\
 &
  \ding{52} &
  90.5 &
  95.6 &
  92.1 &
  92.7 &
  \textbf{92.0} &
  \textbf{82.2} &
  \textbf{39.9} &
  \textbf{40.6} &
  \textbf{26.7} &
  \textbf{56.3} \\ \midrule
\multirow{2}{*}{VisA} &
  &
  82.2 &
  84.9 &
  81.1 &
  82.7 &
  95.4 &
  86.9 &
  20.1 &
  27.2 &
  16.9 &
  49.3 \\
 &
  \ding{52} &
  \textbf{82.5} &
  \textbf{85.2} &
  \textbf{81.6} &
  \textbf{83.1} &
  \textbf{95.8} &
  \textbf{88.8} &
  \textbf{23.4} &
  \textbf{30.6} &
  \textbf{19.1} &
  \textbf{51.5} \\ \midrule
\multirow{2}{*}{MPDD} &
  &
  \textbf{76.0} &
  \textbf{78.9} &
  \textbf{79.5} &
  \textbf{78.1} &
  95.1 &
  86.4 &
  25.1 &
  28.4 &
  17.9 &
  50.6 \\
 &
  \ding{52} &
  74.9 &
  77.5 &
  79.0 &
  77.1 &
  \textbf{95.7} &
  \textbf{87.9} &
  \textbf{25.9} &
  \textbf{29.7} &
  \textbf{18.7} &
  \textbf{51.6} \\ \midrule
\multirow{2}{*}{BTAD} &
  &
  92.7 &
  93.7 &
  89.4 &
  91.9 &
  95.5 &
  81.0 &
  \textbf{50.0} &
  \textbf{52.2} &
  \textbf{36.1} &
  \textbf{63.0} \\
 &
  \ding{52} &
  \textbf{92.9} &
  \textbf{94.8} &
  \textbf{90.0} &
  \textbf{92.6} &
  \textbf{95.6} &
  \textbf{81.4} &
  47.7 &
  50.9 &
  35.6 &
  62.2 \\ \midrule
\multirow{2}{*}{SDD} &
  &
  86.3 &
  83.6 &
  77.0 &
  82.3 &
  89.1 &
  \textbf{65.7} &
  28.8 &
  37.1 &
  22.8 &
  48.7 \\
 &
  \ding{52} &
  \textbf{87.0} &
  \textbf{84.8} &
  \textbf{80.3} &
  \textbf{84.1} &
  \textbf{91.3} &
  65.6 &
  \textbf{40.7} &
  \textbf{43.3} &
  \textbf{27.6} &
  \textbf{53.7} \\ \midrule
\multirow{2}{*}{DAGM} &
  &
  97.7 &
  92.6 &
  \textbf{91.2} &
  93.8 &
  95.5 &
  91.5 &
  58.0 &
  59.5 &
  44.0 &
  69.7 \\
 &
  \ding{52} &
  \textbf{97.8} &
  \textbf{92.7} &
  91.2 &
  \textbf{93.9} &
  \textbf{96.2} &
  \textbf{92.4} &
  \textbf{61.0} &
  \textbf{61.0} &
  \textbf{45.7} &
  \textbf{71.3} \\ \midrule
\multirow{2}{*}{\shortstack{DTD\\-Synthetic}} &
  &
  \textbf{93.5} &
  \textbf{96.9} &
  \textbf{93.7} &
  \textbf{94.7} &
  98.0 &
  93.5 &
  66.7 &
  64.9 &
  48.4 &
  74.3 \\
 &
  \ding{52} &
  93.4 &
  96.8 &
  93.5 &
  94.6 &
  \textbf{98.6} &
  \textbf{94.1} &
  \textbf{72.6} &
  \textbf{67.8} &
  \textbf{51.7} &
  \textbf{76.9} \\ \midrule
\multirow{2}{*}{Average} &
  &
  \textbf{88.4} &
  89.5 &
  86.3 &
  88.1 &
  94.2 &
  83.8 &
  40.1 &
  43.6 &
  29.9 &
  58.3 \\
 &
  \ding{52} &
  88.4 &
  \textbf{89.6} &
  \textbf{86.8} &
  \textbf{88.3} &
  \textbf{95.0} &
  \textbf{84.6} &
  \textbf{44.5} &
  \textbf{46.3} &
  \textbf{32.2} &
  \textbf{60.5} \\ \bottomrule
\end{tabular}
\end{adjustbox}
\end{table}

\begin{table}[!t]
\caption{Ablation study results of LHW in medical domain at pixel level without JAS. Higher performance shown in bold.}
\label{tab:LHW_ablation_medical_pixel}
\centering
\begin{adjustbox}{max width=0.7\textwidth}
\begin{tabular}{@{}cc*{6}{w{c}{1.2cm}}@{}}
\toprule
\multirow{2}{*}{Dataset} &
  \multirow{2}{*}{LHW} &
  \multicolumn{6}{c}{Pixel-level} \\ \cmidrule(lr){3-8}
 &
  &
  ROC &
  PRO &
  AP &
  $\text{F1}_{\text{max}}$ &
  $\text{IoU}_{\text{max}}$ &
  mAD \\ \midrule
\multirow{2}{*}{ISIC} &
  &
  88.8 &
  80.0 &
  76.6 &
  70.9 &
  54.9 &
  74.3 \\
 &
  \ding{52} &
  \textbf{90.1} &
  \textbf{81.7} &
  \textbf{78.8} &
  \textbf{72.1} &
  \textbf{56.4} &
  \textbf{75.8} \\ \midrule
\multirow{2}{*}{CVC-ColonDB} &
  &
  83.7 &
  \textbf{72.8} &
  \textbf{31.6} &
  \textbf{37.8} &
  \textbf{23.3} &
  \textbf{49.8} \\
 &
  \ding{52} &
  \textbf{83.8} &
  72.8 &
  30.3 &
  37.1 &
  22.8 &
  49.4 \\ \midrule
\multirow{2}{*}{CVC-ClinicDB} &
  &
  84.9 &
  70.3 &
  37.7 &
  44.1 &
  28.3 &
  53.1 \\
 &
  \ding{52} &
  \textbf{85.6} &
  \textbf{71.0} &
  \textbf{38.1} &
  \textbf{44.2} &
  \textbf{28.4} &
  \textbf{53.5} \\ \midrule
\multirow{2}{*}{Kvasir} &
  &
  82.0 &
  47.4 &
  42.6 &
  49.9 &
  33.2 &
  51.0 \\
 &
  \ding{52} &
  \textbf{82.8} &
  \textbf{48.0} &
  \textbf{43.2} &
  \textbf{50.0} &
  \textbf{33.3} &
  \textbf{51.5} \\ \midrule
\multirow{2}{*}{Endo} &
  &
  85.8 &
  66.5 &
  48.0 &
  \textbf{52.7} &
  \textbf{35.8} &
  \textbf{57.7} \\
 &
  \ding{52} &
  \textbf{86.1} &
  \textbf{66.8} &
  \textbf{48.1} &
  52.2 &
  35.3 &
  57.7 \\ \midrule
\multirow{2}{*}{TN3K} &
  &
  77.9 &
  44.4 &
  40.8 &
  43.4 &
  27.7 &
  46.8 \\
 &
  \ding{52} &
  \textbf{80.9} &
  \textbf{46.4} &
  \textbf{44.3} &
  \textbf{45.9} &
  \textbf{29.8} &
  \textbf{49.5} \\ \midrule
\multirow{2}{*}{Average} &
  &
  83.8 &
  63.6 &
  46.2 &
  49.8 &
  33.9 &
  55.5 \\
 &
  \ding{52} &
  \textbf{84.9} &
  \textbf{64.4} &
  \textbf{47.2} &
  \textbf{50.2} &
  \textbf{34.3} &
  \textbf{56.2} \\ \bottomrule
\end{tabular}
\end{adjustbox}
\end{table}

\begin{table}[!ht]
\caption{Ablation study results of LHW in medical domain at image level without JAS. Higher performance shown in bold.}
\label{tab:LHW_ablation_medical_image}
\vspace{5pt}
\centering
\begin{adjustbox}{max width=0.5\textwidth}
\begin{tabular}{@{}cc*{4}{w{c}{1.2cm}}@{}}
\toprule
\multirow{2}{*}{Dataset} &
  \multirow{2}{*}{LHW} &
  \multicolumn{4}{c}{Image-level} \\ \cmidrule(lr){3-6}
 &
  &
  ROC &
  AP &
  $\text{F1}_{\text{max}}$ &
  mAD \\ \midrule
\multirow{2}{*}{HeadCT} &
  &
  92.9 &
  92.3 &
  89.4 &
  91.6 \\
 &
  \ding{52} &
  \textbf{93.5} &
  \textbf{93.1} &
  \textbf{90.1} &
  \textbf{92.2} \\ \midrule
\multirow{2}{*}{BrainMRI} &
  &
  93.0 &
  94.3 &
  90.7 &
  92.7 \\
 &
  \ding{52} &
  \textbf{93.0} &
  \textbf{94.4} &
  \textbf{91.8} &
  \textbf{93.1} \\ \midrule
\multirow{2}{*}{Br35h} &
  &
  95.3 &
  95.4 &
  88.2 &
  93.0 \\
 &
  \ding{52} &
  \textbf{95.7} &
  \textbf{95.7} &
  \textbf{89.8} &
  \textbf{93.7} \\ \midrule
\multirow{2}{*}{COVID-19} &
  &
  79.5 &
  \textbf{57.0} &
  \textbf{56.0} &
  64.2 \\
 &
  \ding{52} &
  \textbf{80.7} &
  56.9 &
  55.1 &
  \textbf{64.2} \\ \midrule
\multirow{2}{*}{Average} &
  &
  90.2 &
  84.8 &
  81.1 &
  85.4 \\
 &
  \ding{52} &
  \textbf{90.7} &
  \textbf{85.0} &
  \textbf{81.7} &
  \textbf{85.8} \\ \bottomrule
\end{tabular}
\end{adjustbox}
\end{table}

\subsubsection{Learnable Head Weights Ablation}
The domain adaptation capability of LHW is first validated by comparing performance with and without LHW across pixel-level and image-level detection tasks. In the industrial domain, as shown in Table~\ref{tab:LHW_ablation_industrial}, LHW leads to a 2.2\%p improvement in pixel-level mAD, with 88.6\% (31 out of 35) of evaluation cases showing performance gains. Similarly, in the medical domain, Table~\ref{tab:LHW_ablation_medical_pixel} reports a 0.8\%p increase in mAD, with 80\% (24 out of 30) of cases improving.

Image-level detection results further reinforce LHW's effectiveness. In the industrial domain, LHW yields consistent mAD gains across all datasets. In the medical domain (Table~\ref{tab:LHW_ablation_medical_image}), performance improves across all datasets and all evaluation metrics. Notably, LHW is not directly involved in image-level score computation, but its domain adaptation effects appear to influence the shared text embedding through gradient flow from the anomaly map. This observation suggests that LHW contributes not only to local feature alignment but also enhances global semantic understanding, resulting in improved performance at both levels.

\begin{table}[!t]
\caption{Ablation study results of LHW under different local feature extraction settings, showing average mAD of HeadCLIP without JAS in industrial and medical domains. Each setting corresponds to the use of $F_{local}$, obtained by aggregating features from the specified layers in the image encoder's local path. Higher performance is shown in bold.}
\label{tab:Feature_LHW_ablation}
\centering
\begin{adjustbox}{max width=0.8\textwidth}
\begin{tabular}{@{}cc*{4}{w{c}{2.2cm}}@{}}
\toprule
\multirow{2}{*}{Layer Index} &
  \multirow{2}{*}{LHW} &
  \multicolumn{2}{c}{Industrial} &
  \multicolumn{2}{c}{Medical} \\ \cmidrule(lr){3-4} \cmidrule(lr){5-6}
 &
  &
  Image-mAD &
  Pixel-mAD &
  Image-mAD &
  Pixel-mAD \\ \midrule
\multirow{2}{*}{6, 12, 18, 24} &
  &
  86.0 &
  \textbf{84.1} &
  \textbf{55.6} &
  53.0 \\
 &
  \ding{52} &
  \textbf{87.7} &
  84.0 &
  54.8 &
  \textbf{54.8} \\ \midrule
\multirow{2}{*}{12, 18, 24} &
  &
  \textbf{86.9} &
  \textbf{84.2} &
  58.3 &
  53.4 \\
 &
  \ding{52} &
  86.7 &
  83.1 &
  \textbf{58.7} &
  \textbf{54.7} \\ \midrule
\multirow{2}{*}{18, 24} &
  &
  87.6 &
  84.9 &
  59.5 &
  52.9 \\
 &
  \ding{52} &
  \textbf{87.8} &
  \textbf{85.3} &
  \textbf{60.8} &
  \textbf{54.9} \\ \midrule
\multirow{2}{*}{24} &
  &
  87.9 &
  86.1 &
  58.4 &
  54.0 \\
 &
  \ding{52} &
  \textbf{88.1} &
  \textbf{86.2} &
  \textbf{60.3} &
  \textbf{57.4} \\ \midrule
\multirow{2}{*}{18} &
  &
  87.8 &
  \textbf{86.0} &
  57.3 &
  50.3 \\
 &
  \ding{52} &
  \textbf{88.0} &
  85.4 &
  \textbf{59.7} &
  \textbf{52.9} \\ \midrule
\multirow{2}{*}{12} &
  &
  86.8 &
  \textbf{83.7} &
  45.0 &
  49.8 \\
 &
  \ding{52} &
  \textbf{86.9} &
  83.1 &
  \textbf{49.2} &
  \textbf{52.0} \\ \midrule
\multirow{2}{*}{6} &
  &
  86.6 &
  89.1 &
  32.3 &
  47.7 \\
 &
  \ding{52} &
  \textbf{86.8} &
  \textbf{90.2} &
  \textbf{34.1} &
  \textbf{48.1} \\ \bottomrule
\end{tabular}
\end{adjustbox}
\end{table}

\subsubsection{Learnable Head Weights Across Local Feature Extraction Settings}
Whether the effectiveness of LHW depends on the choice of transformer layers used for local feature extraction is next investigated. Multiple configurations are tested by varying the layers from which $F_{\text{local}}$ is aggregated.

In the industrial domain, LHW consistently improves image-level mAD in all seven layer configurations, with gains ranging from 0.2 to 1.7\%p. Pixel-level performance also improves in most configurations. A similar pattern is observed in the medical domain, where LHW yields image-level improvements of 1.3 to 4.2\%p and pixel-level improvements of 0.4 to 3.4\%p across different settings.

These results demonstrate that the proposed LHW module contributes robustly across diverse architectural configurations, further validating its general applicability for anomaly detection in varying domain scenarios.

\begin{table}[!t]
\caption{Ablation study results of initialization value of LHW, showing average mAD of HeadCLIP in industrial and medical domains. Best performance shown in bold, second-best underlined.}
\label{tab:LHW_initialization_ablation}
\centering
\begin{adjustbox}{max width=0.5\textwidth}
\begin{tabular}{@{}c*{2}{w{c}{1.5cm}}@{}}
\toprule
\multirow{2}{*}{Initialization} &
  \multicolumn{2}{c}{Pixel-mAD} \\ \cmidrule(lr){2-3}
 &
  Industrial &
  Medical \\ \midrule
Gaussian &
  59.6 &
  54.5 \\
0.0 &
  60.1 &
  54.6 \\
0.5 &
  {\ul 60.4} &
  {\ul 55.8} \\
1.0 &
  \textbf{60.5} &
  \textbf{56.2} \\ \bottomrule
\end{tabular}
\end{adjustbox}
\end{table}

\subsubsection{Learnable Head Weights Initialization Value Ablation}
To evaluate the sensitivity of LHW to different initialization strategies, the effects of Gaussian and uniform initialization methods on pixel-level anomaly detection performance are compared. Table~\ref{tab:LHW_initialization_ablation} summarizes the results across both industrial and medical domains.

\paragraph{Gaussian initialization} When LHW is initialized with Gaussian-distributed random values, performance significantly deteriorates, achieving 59.6\% mAD in the industrial domain and 54.5\% in the medical domain. This result suggests that random initial weights fail to reflect the functional characteristics of individual attention heads, leading to suboptimal or unstable domain adaptation during training.
\paragraph{Uniform initialization} Uniform initializations with values of 0.0, 0.5, and 1.0 are further investigated. A gradual improvement is observed as the initialization value increases. In the industrial domain, pixel-level performance rises from 60.1\% (init=0.0) to 60.5\% (init=1.0). In the medical domain, performance improves from 54.6\% to 56.2\% across the same initialization range.

These results indicate that initializing LHW with uniform values closer to 1.0, which maintains balanced attention contribution at the start of training, facilitates more effective learning. In contrast, assigning arbitrary or near-zero weights appears to hinder the model's ability to leverage the expressive power of individual heads, especially in early training stages.

\begin{table}[!t]
\caption{Comparison of amplified and suppressed attention heads based on TextSpan analysis.}
\label{tab:LHW_textspan}
\centering
\scriptsize
\begin{tabularx}{\textwidth}{@{}lXX@{}}
\toprule
\textbf{Aspect} & \textbf{Amplified Heads (w$>$1)} & \textbf{Suppressed Heads (w$\leq$1)} \\ 
\midrule
Color Encoding & Full spectrum, consistent specialization & Partial, mixed with geographic noise \\
\midrule
Texture Features & Fine-grained patterns, surface details & Absent or overwhelmed by noise \\
\midrule
Spatial Structure & Scene layout, geometric regularity & Generic location terms \\
\midrule
Scale/Measurement & Size, dimensions, proportions & Abstract terms like ``overall'' \\
\midrule
Geographic Terms & Absent & Dominant (korea, scotland, connecticut, washington, missouri, etc.) \\
\midrule
Web/Commerce Noise & Absent & Dominant (social, wildlife, enable, rom, psp, cvs, msn, aol, ebay, tripadvisor) \\
\midrule
Brand Names & Absent & Present (motorola, panasonic, nokia, ericsson, playstation) \\
\midrule
Abstract Functional Terms & Absent & Present (overall, enable, initial, indicated, conducted, whether) \\
\bottomrule
\end{tabularx}
\end{table}

\subsubsection{Qualitative Analysis of Learned Head Weights}

To understand what visual concepts the learned head weights prioritize, the semantic specialization of attention heads is analyzed using TextSpan~\citep{textspan}. Table~\ref{tab:LHW_textspan} summarizes the key differences between amplified (w$>$1) and suppressed (w$\leq$1) heads.

Amplified heads consistently specialize in anomaly-relevant features: full-spectrum color encoding for detecting chromatic deviations, fine-grained texture patterns for surface defect identification, spatial structure for geometric irregularity detection, and scale/measurement attributes for dimensional assessment. In contrast, suppressed heads are dominated by task-irrelevant features including geographic locations (korea, scotland, washington), web/commerce terminology (social, wildlife, psp, cvs), brand names (motorola, panasonic, nokia), and abstract functional terms (overall, enable, initial).

This systematic filtering demonstrates that the learnable head weights effectively distinguish between anomaly-relevant visual features and noisy web-crawled artifacts inherited from CLIP's pre-training corpus. The learned weights thus serve as an automatic feature selection mechanism, amplifying texture and color sensitivity while suppressing irrelevant semantic noise without requiring manual head selection. Detailed layer-by-layer analysis of individual attention heads, including the hierarchical progression of feature specialization from early to late layers, is provided in Appendix~\ref{appendix:textspan_analysis}.

\subsection{Additional Study}

\subsubsection{Computational Efficiency Comparison}

In addition to performance gains, a key requirement for real-world anomaly detection is computational efficiency. While HeadCLIP introduces two new components, both are designed to be lightweight and practical for deployment. In particular, LHW adds only 240 trainable scalar parameters in total, which is negligible relative to the 427M parameters of the frozen CLIP backbone. Furthermore, JAS is applied only during inference, and involves a simple weighted fusion of precomputed scores without any additional forward passes or memory overhead.

To verify that these enhancements do not compromise efficiency, the total parameter count, inference latency, and throughput are compared with AdaCLIP and AnomalyCLIP under identical backbone settings (ViT-L/14@336px).

\begin{table}[!t]
\caption{Computational efficiency comparison of different anomaly detection methods. All models use ViT-L/14@336px as the backbone architecture. Inference time and throughput are measured per image, averaged over 50 runs on a single NVIDIA RTX 2080 GPU.}
\label{tab:efficiency_comparison}
\centering
\scriptsize
\begin{adjustbox}{max width=\textwidth}
\begin{tabular}{@{}l*{4}{w{c}{2.5cm}}@{}}
\toprule
Method &
  Total Params &
  Additional Params &
  Inference Time (ms) &
  Throughput (FPS) \\ \midrule
AdaCLIP &
  439.42M &
  +10.67M &
  225.53 &
  4.43 \\
AnomalyCLIP &
  433.50M &
  +5.56M &
  178.95 &
  5.59 \\
HeadCLIP &
  433.50M &
  +5.56M &
  178.63 &
  5.60 \\ \bottomrule
\end{tabular}
\end{adjustbox}
\end{table}

As shown in Table~\ref{tab:efficiency_comparison}, HeadCLIP achieves comparable or even slightly better inference speed and throughput compared to AnomalyCLIP, despite incorporating additional adaptation mechanisms. Its added parameter count remains the same, indicating that LHW introduce virtually no overhead. Moreover, HeadCLIP is significantly more efficient than AdaCLIP, which requires double the parameter addition and suffers from a longer inference time.

These results confirm that the proposed LHW and JAS modules enhance detection performance without degrading computational efficiency. Inference time was measured as the average latency over 50 images, evaluated on a single NVIDIA RTX 2080 GPU after a short warm-up.

\subsubsection{Joint Anomaly Score Ablation}
\begin{figure}[t!]
    \centering
    \includegraphics[width=1\linewidth]{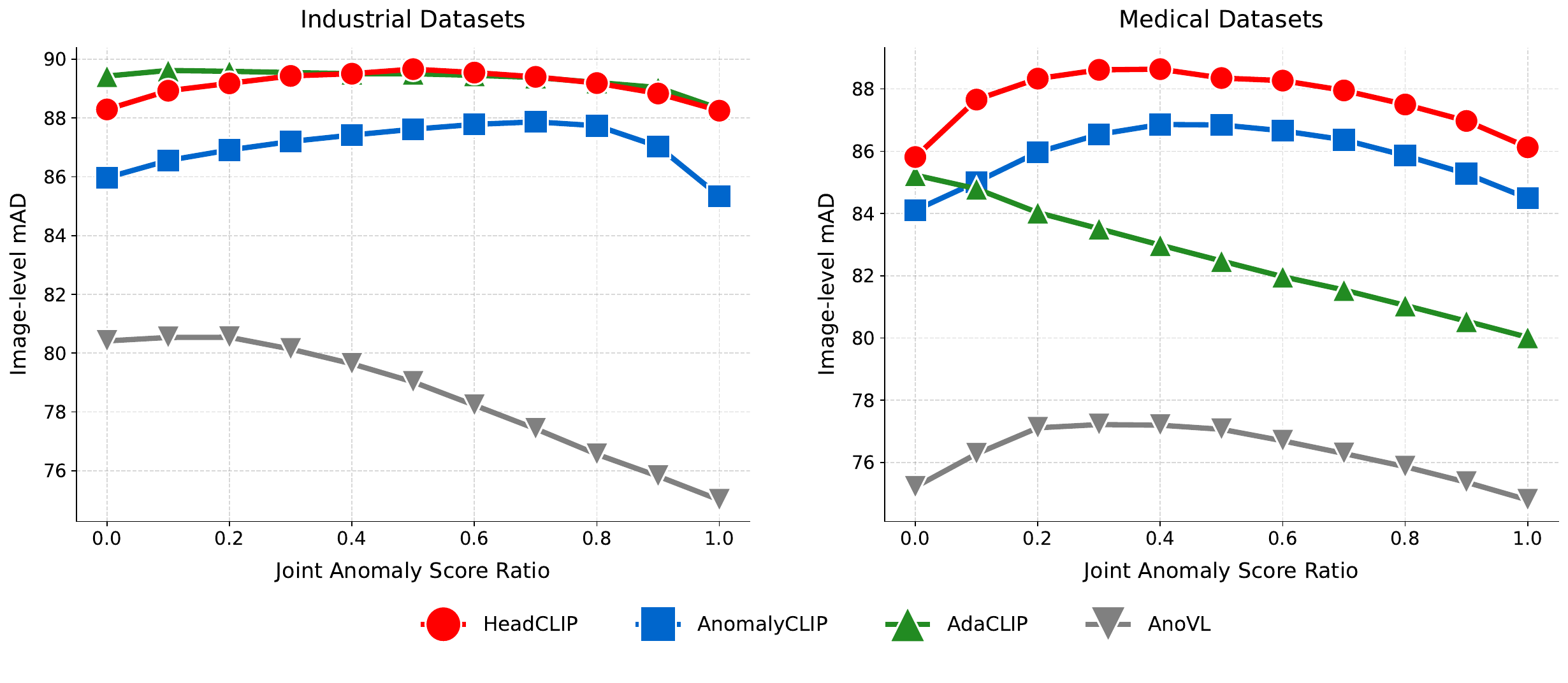}
    \vspace{-30pt}
    \caption{Image-level mAD performance as a function of the JAS ratio (x-axis) on industrial (left) and medical (right) datasets. The plots compare four methods with the y-axis showing image-level mAD. The results illustrate how varying the JAS ratio influences the overall detection performance for each approach.}
    \label{fig:JAS_ablation}
\end{figure}

To evaluate the effectiveness and robustness of the proposed JAS, ablation experiments are conducted by varying the score ratio parameter 
$r$, which controls the relative contribution of global and local anomaly scores. Figure~\ref{fig:JAS_ablation} presents the results across datasets in both industrial and medical domains.

The proposed HeadCLIP framework consistently outperforms existing methods across a wide range of $r$ values. In particular, in the medical domain, HeadCLIP maintains the highest mAD performance across all tested ratio values. In the industrial domain, it achieves either the top or near-top performance in most configurations, indicating its robustness to changes in the score fusion strategy.

Among baseline methods, AnomalyCLIP exhibits relatively stable performance across different ratio values but underperforms HeadCLIP consistently. AdaCLIP shows a notable performance decline as $r$ increases in the medical domain, suggesting poor integration of global-local cues when domain adaptation is suboptimal. AnoVL, which lacks an explicit domain adaptation mechanism, yields the lowest performance in both domains regardless of ratio value.

These findings suggest that JAS is most effective when built upon strong domain-adapted features from both modalities. Moreover, HeadCLIP demonstrates high robustness to the joint score ratio, maintaining stable and superior detection performance across varying fusion strategies and domain characteristics.

\subsubsection{Joint Anomaly Score Top-k Ablation}

\begin{table}[]
\caption{Ablation study on top-k ratio selection for JAS. Values represent mean ($\pm$standard deviation) across 7 industrial and 4 medical datasets. The analysis demonstrates that $k = 0.05$ provides the optimal trade-off between detection performance and stability across different domains.}
\label{tab:Topk_ablation}
\centering
\begin{adjustbox}{max width=0.8\textwidth}
\begin{tabular}{@{}cc*{3}{w{c}{2.5cm}}@{}}
\toprule
Domain &
  Top-k &
  ROC &
  AP &
  $\text{F1}_{\text{max}}$ \\ \midrule
\multirow{3}{*}{Industrial} &
  0.01 &
  89.5 ($\pm$7.4) &
  89.2 ($\pm$8.4) &
  86.1 ($\pm$8.1) \\
 &
  0.05 &
  88.7 ($\pm$7.0) &
  88.8 ($\pm$8.0) &
  85.7 ($\pm$7.5) \\
 &
  0.1 &
  88.0 ($\pm$7.0) &
  88.1 ($\pm$8.0) &
  85.0 ($\pm$7.7) \\ \midrule
\multirow{3}{*}{Medical} &
  0.01 &
  92.3 ($\pm$4.7) &
  85.5 ($\pm$17.7) &
  80.5 ($\pm$15.8) \\
 &
  0.05 &
  92.2 ($\pm$3.0) &
  87.5 ($\pm$12.9) &
  80.7 ($\pm$12.1) \\
 &
  0.1 &
  91.6 ($\pm$2.4) &
  87.5 ($\pm$10.6) &
  80.4 ($\pm$10.0) \\ \bottomrule
\end{tabular}
\end{adjustbox}
\end{table}

An ablation study is conducted to determine the optimal top-k ratio for JAS by evaluating image-level anomaly detection using only the mean of top-k pixel-level anomaly scores across 7 industrial and 4 medical datasets.

Table~\ref{tab:Topk_ablation} shows that while $k = 0.01$ achieves the highest performance in industrial domains, it exhibits substantial variance across datasets. In contrast, $k = 0.05$ offers a better trade-off between performance and stability, maintaining competitive accuracy with consistently lower standard deviations. In medical domains, $k = 0.05$ even outperforms $k = 0.01$ in terms of mAP and F1-max, while significantly reducing variance. Although $k = 0.1$ yields the lowest variance, it consistently underperforms, suggesting that it may include excessive noise.

The value $k = 0.05$ is adopted as the default configuration, as it provides optimal balance between detection performance and cross-dataset stability across both domains.

\subsubsection{Module Ablation}
To assess the individual and combined contributions of the two core components, LHW and the JAS, a module-wise ablation study is conducted. Table~\ref{tab:Total_ablation} summarizes the performance changes when each module is removed from the full HeadCLIP model.

\paragraph{Impact of removing JAS}\label{sec:jas_ablation} Removing JAS leads to a noticeable decline in image-level anomaly detection performance, with a decrease of 1.4\%p in the industrial domain and 2.6\%p in the medical domain. This highlights the importance of leveraging pixel-level domain-adapted features to enhance global anomaly scoring.

\paragraph{Impact of removing both JAS and LHW} When LHW is additionally removed, a further degradation is observed, particularly in pixel-level performance. Specifically, mAD drops by 2.2\%p in the industrial domain and 0.7\%p in the medical domain. This indicates that LHW plays a key role in improving anomaly localization by enabling per-head visual adaptation to domain-specific characteristics.

\paragraph{Summary} The best performance across both image-level and pixel-level detection is achieved when both LHW and JAS are included, confirming that these modules contribute complementary capabilities: LHW enhances fine-grained visual representation, while JAS enables robust score fusion across modalities and scales. Their joint application leads to consistent performance gains in both domains.

\begin{table}[]
\caption{Component ablation study results of LHW and JAS, showing average mAD in industrial and medical domains at both image and pixel levels. Numbers in parentheses indicate the difference in percentage points (\%p).}
\label{tab:Total_ablation}
\centering
\begin{adjustbox}{max width=0.8\textwidth}
\begin{tabular}{@{}ccc*{4}{w{c}{1.5cm}}@{}}
\toprule
\multirow{2}{*}{Model} &
  \multirow{2}{*}{LHW} &
  \multirow{2}{*}{JAS} &
  \multicolumn{2}{c}{Image-level} &
  \multicolumn{2}{c}{Pixel-level} \\ \cmidrule(lr){4-5} \cmidrule(lr){6-7}
 &
  &
  &
  Industrial &
  Medical &
  Industrial &
  Medical \\ \midrule
\multirow{3}{*}{HeadCLIP} &
  \ding{52} &
  \ding{52} &
  89.7 &
  88.4 &
  60.5 &
  56.2 \\
 &
  \ding{52} &
  &
  88.3\textcolor{blue}{(-1.4)} &
  85.8\textcolor{blue}{(-2.6)} &
  60.5 &
  56.2 \\
 &
  &
  &
  88.1\textcolor{blue}{(-0.2)} &
  85.4\textcolor{blue}{(-0.4)} &
  58.3\textcolor{blue}{(-2.2)} &
  55.5\textcolor{blue}{(-0.7)} \\ \bottomrule
\end{tabular}
\end{adjustbox}
\end{table}

\section{Conclusion}\label{Conclusion}

This paper presents Head-adaptive CLIP (HeadCLIP), a novel ZSAD framework that enables effective domain adaptation of both text and image encoders. The proposed learnable head weights (LHW) exploit the semantic specialization of attention heads in Vision Transformers, achieving domain-specific adaptation with only 304 learnable parameters. Combined with the joint anomaly score (JAS) that integrates pixel-level and image-level evidence, HeadCLIP consistently outperforms prior ZSAD methods across 17 datasets in industrial and medical domains, with improvements of up to 4.9\%p in pixel-level mAD and 3.7\%p in image-level mAD.

Despite these contributions, several limitations remain. First, when trained exclusively on MVTec AD, HeadCLIP exhibits reduced performance on multi-object scenes and visually complex environments, such as the capsules, candles, and pcb2 categories in VisA. This suggests that the learned head weights may encode inductive biases toward single-object, well-centered anomaly patterns. Second, as a semantic prior-based approach, HeadCLIP's anomaly decisions are inherently tied to the visual concepts learned during training, which may limit adaptability when target domains exhibit substantially different characteristics.

These limitations motivate several future research directions: (1) investigating joint training strategies that incorporate diverse scene compositions, including multi-object and cluttered environments, to learn more generalizable head weight configurations; (2) developing adaptive head weighting mechanisms that dynamically adjust to target domain characteristics during inference without additional training; and (3) exploring integration with open-set recognition and continual learning paradigms to enable incremental adaptation to emerging anomaly types.

\bibliographystyle{elsarticle-harv}
\bibliography{reference}
\clearpage

\section{Appendix}\label{sec:appendix}

\subsection{Failure Cases}
\begin{figure}[!ht]
    \centering
    \includegraphics[width=1\linewidth]{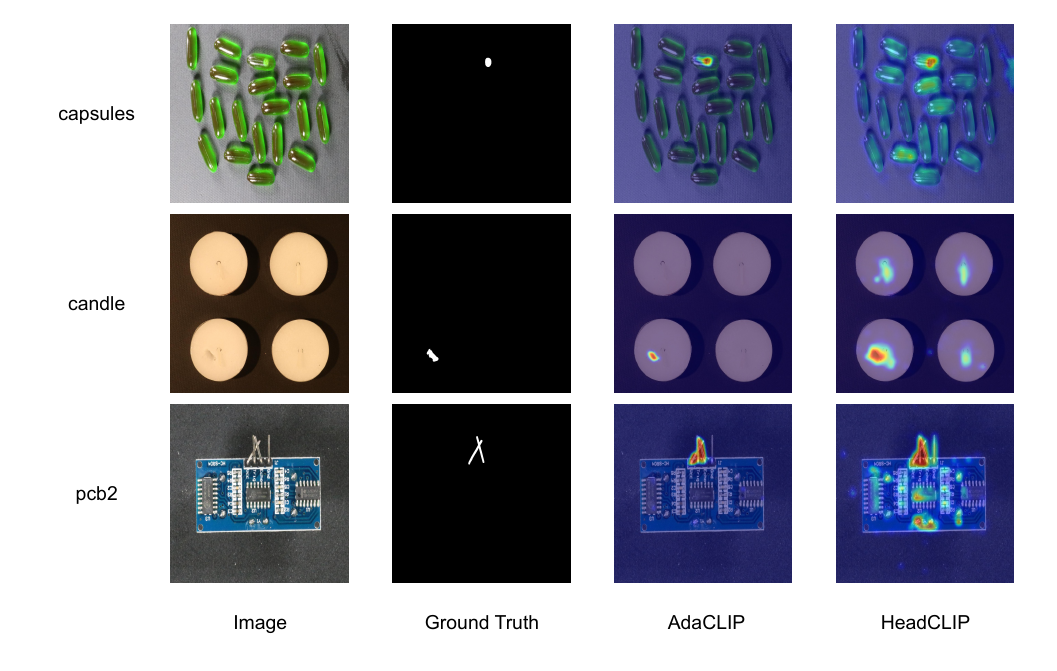}
    \vspace{-30pt}
    \caption{Representative failure cases from VisA dataset illustrating HeadCLIP's limitations compared to AdaCLIP. HeadCLIP struggles with anomalies that deviate from MVTecAD training patterns: missing defects in multi-object capsules, weak localization on irregular candle surfaces, and reduced sensitivity to fine-grained pcb2 circuit defects. These cases highlight the model's inductive bias toward canonical single-object anomaly scenarios.}
    \label{fig:failures}
\end{figure}

Qualitative analyses of failure cases observed during inference are provided to enhance the transparency and fairness of the evaluation. HeadCLIP demonstrates consistently strong performance across a wide range of datasets and metrics. This strong performance can be attributed to the proposed LHW and the JAS. However, the model does not always achieve top accuracy for every individual subclass.

To provide fair evaluation and better understand HeadCLIP's characteristics, representative examples where the model struggled to detect or localize anomalies effectively are included. These cases were selected from both industrial and medical domains and reflect settings where anomaly characteristics deviate from those seen during training.

As illustrated in Figure~\ref{fig:failures} and Table~\ref{tab:appendix_visa}, HeadCLIP exhibits relatively lower performance in scenarios involving multiple object instances, such as the capsules and candles classes, or scenes with high visual complexity, as seen in the pcb2 class, particularly compared to AdaCLIP. In contrast, the model shows superior performance on canonical single-object anomalies.

It is hypothesized that this behavior results from the adaptation dynamics introduced by LHW. Specifically, when trained on MVTecAD, which consists exclusively of images containing a single, well-centered object, HeadCLIP's LHW are implicitly optimized to favor anomaly patterns common in such settings. Consequently, when evaluated on datasets like VisA that include multiple objects or cluttered visual environments, this inductive bias may limit the model's ability to handle visually irregular or complex scenes.

HeadCLIP consistently achieves the highest average performance across all datasets, confirming its robustness in diverse domains. Nevertheless, the model has limitations, particularly in detecting anomalies that significantly deviate from the patterns observed in training data. Addressing these challenges and further improving generalization to such scenarios represents an important direction for future work.

\subsection{Detailed Subclass-level Results}
This section provides comprehensive subclass-level experimental results to offer a more granular analysis of HeadCLIP's performance across different anomaly types. Performance metrics are reported at both image and pixel levels, enabling detailed examination of the model's effectiveness and consistency across diverse domains.

It should be noted that only datasets containing multiple subclasses are included in this section. Datasets comprising a single subclass, specifically, SDD in the industrial domain, and all medical domain datasets (ISIC, CVC-ClinicDB, CVC-ColonDB, Kvasir, Endo, TN3k, HeadCT, BrainMRI, Br35H, and COVID-19), are omitted, as their subclass-level results are equivalent to the dataset-level results presented in the main text.

Tables \ref{tab:appendix_mvtec}--\ref{tab:appendix_dtd} present the detailed subclass-level performance for the MVTecAD, VisA, MPDD, BTAD, DAGM, and DTD-Synthetic datasets, respectively.

\renewcommand{\arraystretch}{0.5}  % 행간 간격 축소

\begin{longtable}{@{}>{\scriptsize}c>{\scriptsize}l*{8}{>{\scriptsize\centering\arraybackslash}p{0.8cm}}@{}}
\caption{ZSAD performance comparison on MVTecAD. Best performance shown in \textbf{bold}, second-best shown in {\ul underline}.}
\label{tab:appendix_mvtec} \\
\toprule
\multirow{2}{*}{\scriptsize Subclass} & \multirow{2}{*}{\scriptsize Model} & \multicolumn{3}{c}{\scriptsize Image-level} & \multicolumn{5}{c}{\scriptsize Pixel-level} \\ 
\cmidrule(lr){3-5} \cmidrule(lr){6-10}
 & & ROC & AP & $\text{F1}_{\text{max}}$ & ROC & PRO & AP & $\text{F1}_{\text{max}}$ & $\text{IoU}_{\text{max}}$ \\ 
\midrule
\endfirsthead

\multicolumn{10}{c}{\tablename\ \thetable{} -- continued from previous page} \\
\toprule
\multirow{2}{*}{\scriptsize Subclass} & \multirow{2}{*}{\scriptsize Model} & \multicolumn{3}{c}{\scriptsize Image-level} & \multicolumn{5}{c}{\scriptsize Pixel-level} \\ 
\cmidrule(lr){3-5} \cmidrule(lr){6-10}
 & & ROC & AP & $\text{F1}_{\text{max}}$ & ROC & PRO & AP & $\text{F1}_{\text{max}}$ & $\text{IoU}_{\text{max}}$ \\ 
\midrule
\endhead

\multicolumn{10}{r}{\scriptsize Continued on next page} \\
\endfoot

\bottomrule
\endlastfoot

\multirow{4}{*}{bottle} 
 & AnoVL & \textbf{97.5} & \textbf{99.3} & \textbf{96.8} & \textbf{92.1} & {\ul 77.9} & {\ul 56.0} & \textbf{53.6} & \textbf{36.6} \\
 & AnomalyCLIP & 84.0 & 95.3 & 86.4 & 86.1 & 76.9 & 49.1 & 47.9 & 31.5 \\
 & AdaCLIP & {\ul 97.3} & {\ul 99.2} & {\ul 95.3} & {\ul 91.8} & 38.7 & \textbf{61.8} & 33.0 & 19.7 \\
 & HeadCLIP & 93.8 & 98.3 & 92.6 & 90.8 & \textbf{80.8} & 54.9 & {\ul 51.4} & {\ul 34.6} \\ 
\midrule
\multirow{4}{*}{cable} 
 & AnoVL & \textbf{86.1} & \textbf{90.8} & \textbf{84.3} & 70.7 & \textbf{69.1} & 17.9 & {\ul 27.0} & {\ul 15.6} \\
 & AnomalyCLIP & 65.2 & 77.6 & 77.9 & 67.2 & 51.4 & 5.6 & 10.8 & 5.7 \\
 & AdaCLIP & 76.6 & 86.1 & 79.3 & {\ul 82.2} & 43.3 & \textbf{24.6} & \textbf{28.8} & \textbf{16.8} \\
 & HeadCLIP & {\ul 80.1} & {\ul 87.6} & {\ul 81.3} & \textbf{84.0} & {\ul 67.9} & {\ul 17.9} & 23.7 & 13.4 \\ 
\midrule
\multirow{4}{*}{capsule} 
 & AnoVL & 70.9 & 91.6 & {\ul 93.1} & 86.9 & 59.9 & 8.4 & 15.2 & 8.2 \\
 & AnomalyCLIP & 86.4 & 97.1 & 91.0 & 93.4 & \textbf{89.5} & 24.7 & {\ul 28.6} & {\ul 16.7} \\
 & AdaCLIP & \textbf{93.0} & \textbf{98.5} & \textbf{93.3} & \textbf{96.4} & 27.8 & \textbf{38.7} & 19.6 & 10.9 \\
 & HeadCLIP & {\ul 88.6} & {\ul 97.6} & 92.3 & {\ul 95.1} & {\ul 88.8} & {\ul 26.4} & \textbf{30.2} & \textbf{17.8} \\ 
\midrule
\multirow{4}{*}{carpet} 
 & AnoVL & 99.5 & 99.8 & 98.3 & 97.8 & 89.6 & 45.0 & 49.1 & 32.5 \\
 & AnomalyCLIP & {\ul 100.0} & {\ul 100.0} & {\ul 99.4} & 98.8 & \textbf{96.7} & {\ul 63.5} & \textbf{63.1} & \textbf{46.1} \\
 & AdaCLIP & \textbf{100.0} & \textbf{100.0} & \textbf{100.0} & {\ul 98.9} & 55.9 & \textbf{65.1} & 26.7 & 15.4 \\
 & HeadCLIP & \textbf{100.0} & \textbf{100.0} & \textbf{100.0} & \textbf{98.9} & {\ul 93.7} & 63.0 & {\ul 61.7} & {\ul 44.6} \\ 
\midrule
\multirow{4}{*}{grid} 
 & AnoVL & \textbf{99.9} & \textbf{100.0} & \textbf{99.1} & {\ul 96.1} & \textbf{88.5} & 15.2 & 23.3 & 13.2 \\
 & AnomalyCLIP & 96.2 & 98.7 & {\ul 94.4} & 93.2 & {\ul 79.7} & 23.3 & {\ul 32.4} & {\ul 19.3} \\
 & AdaCLIP & {\ul 99.8} & {\ul 99.9} & \textbf{99.1} & 92.4 & 3.2 & {\ul 29.6} & 5.0 & 2.6 \\
 & HeadCLIP & 96.9 & 99.0 & {\ul 94.4} & \textbf{97.8} & 58.9 & \textbf{31.5} & \textbf{35.3} & \textbf{21.4} \\ 
\midrule
\multirow{4}{*}{hazelnut} 
 & AnoVL & 91.4 & 95.6 & 87.9 & 94.7 & 78.5 & 29.6 & 36.0 & 22.0 \\
 & AnomalyCLIP & {\ul 95.2} & \textbf{97.7} & \textbf{92.9} & {\ul 96.7} & \textbf{94.8} & {\ul 48.4} & {\ul 49.5} & {\ul 32.9} \\
 & AdaCLIP & \textbf{95.3} & {\ul 97.5} & {\ul 92.1} & \textbf{97.6} & 67.2 & \textbf{57.1} & \textbf{54.0} & \textbf{37.0} \\
 & HeadCLIP & 91.0 & 95.3 & 88.3 & 96.6 & {\ul 89.6} & 46.1 & 46.4 & 30.2 \\ 
\midrule
\multirow{4}{*}{leather} 
 & AnoVL & \textbf{100.0} & \textbf{100.0} & \textbf{100.0} & 98.9 & {\ul 95.3} & 33.8 & 37.3 & 22.9 \\
 & AnomalyCLIP & {\ul 99.8} & {\ul 99.9} & 98.9 & 99.0 & \textbf{97.1} & 35.3 & 37.6 & 23.1 \\
 & AdaCLIP & 99.8 & 99.9 & {\ul 99.5} & {\ul 99.3} & 53.7 & \textbf{57.4} & {\ul 38.9} & {\ul 24.2} \\
 & HeadCLIP & \textbf{100.0} & \textbf{100.0} & \textbf{100.0} & \textbf{99.3} & 94.6 & {\ul 47.1} & \textbf{48.1} & \textbf{31.6} \\ 
\midrule
\multirow{4}{*}{metal nut} 
 & AnoVL & {\ul 93.3} & {\ul 98.5} & {\ul 93.0} & 71.4 & 52.5 & 21.6 & 29.7 & 17.5 \\
 & AnomalyCLIP & 90.7 & 97.7 & 92.9 & \textbf{80.2} & {\ul 72.7} & {\ul 30.9} & \textbf{36.7} & \textbf{22.5} \\
 & AdaCLIP & 78.4 & 95.3 & 89.4 & 68.6 & 39.5 & 22.1 & 21.0 & 11.7 \\
 & HeadCLIP & \textbf{95.5} & \textbf{98.9} & \textbf{96.3} & {\ul 78.8} & \textbf{75.9} & \textbf{32.1} & {\ul 35.2} & {\ul 21.3} \\ 
\midrule
\multirow{4}{*}{pill} 
 & AnoVL & 83.8 & 96.6 & 91.6 & 80.8 & 72.5 & 15.0 & 20.5 & 11.4 \\
 & AnomalyCLIP & 77.5 & 94.0 & 92.4 & {\ul 89.6} & {\ul 90.2} & {\ul 26.4} & {\ul 29.4} & {\ul 17.2} \\
 & AdaCLIP & \textbf{91.8} & \textbf{98.3} & \textbf{94.7} & \textbf{91.7} & 35.5 & \textbf{33.8} & 16.9 & 9.2 \\
 & HeadCLIP & {\ul 85.4} & {\ul 96.7} & {\ul 93.3} & 85.9 & \textbf{91.4} & 26.3 & \textbf{29.7} & \textbf{17.4} \\ 
\midrule
\multirow{4}{*}{screw} 
 & AnoVL & 76.4 & 91.2 & 86.3 & 91.3 & 64.9 & 5.8 & 13.5 & 7.2 \\
 & AnomalyCLIP & {\ul 87.8} & \textbf{95.6} & {\ul 89.2} & {\ul 97.7} & \textbf{91.3} & {\ul 25.7} & {\ul 29.7} & {\ul 17.4} \\
 & AdaCLIP & \textbf{89.1} & {\ul 95.6} & \textbf{91.9} & \textbf{98.7} & 63.3 & \textbf{37.0} & \textbf{41.9} & \textbf{26.5} \\
 & HeadCLIP & 82.0 & 93.0 & 88.8 & 96.6 & {\ul 84.9} & 19.8 & 25.0 & 14.3 \\ 
\midrule
\multirow{4}{*}{tile} 
 & AnoVL & {\ul 100.0} & {\ul 100.0} & {\ul 99.4} & 91.2 & 74.2 & 43.4 & 49.8 & 33.1 \\
 & AnomalyCLIP & \textbf{100.0} & \textbf{100.0} & \textbf{100.0} & {\ul 95.0} & \textbf{89.6} & {\ul 65.1} & {\ul 64.7} & {\ul 47.8} \\
 & AdaCLIP & 99.7 & 99.9 & 98.8 & 88.6 & 11.3 & 59.4 & 13.6 & 7.3 \\
 & HeadCLIP & 99.6 & 99.9 & 98.8 & \textbf{96.8} & {\ul 89.0} & \textbf{72.1} & \textbf{67.8} & \textbf{51.3} \\ 
\midrule
\multirow{4}{*}{toothbrush} 
 & AnoVL & {\ul 93.9} & {\ul 97.7} & {\ul 92.1} & 92.6 & {\ul 81.2} & 13.3 & 23.5 & 13.3 \\
 & AnomalyCLIP & 83.3 & 92.0 & {\ul 92.1} & 90.0 & 81.1 & 10.4 & 16.8 & 9.2 \\
 & AdaCLIP & 90.8 & 96.6 & 89.6 & \textbf{97.2} & 59.3 & \textbf{45.8} & \textbf{45.7} & \textbf{29.7} \\
 & HeadCLIP & \textbf{99.2} & \textbf{99.7} & \textbf{98.3} & {\ul 95.6} & \textbf{93.5} & {\ul 36.1} & {\ul 39.8} & {\ul 24.8} \\ 
\midrule
\multirow{4}{*}{transistor} 
 & AnoVL & 88.5 & {\ul 87.9} & 82.1 & \textbf{73.8} & {\ul 54.9} & \textbf{22.0} & \textbf{30.2} & \textbf{17.8} \\
 & AnomalyCLIP & \textbf{92.8} & \textbf{91.5} & {\ul 82.2} & 65.1 & \textbf{54.9} & 13.1 & 16.4 & 8.9 \\
 & AdaCLIP & 86.2 & 87.1 & 77.8 & 63.3 & 24.8 & 14.7 & 9.2 & 4.8 \\
 & HeadCLIP & {\ul 89.9} & 86.7 & \textbf{82.2} & {\ul 70.7} & 54.6 & {\ul 15.5} & {\ul 16.9} & {\ul 9.2} \\ 
\midrule
\multirow{4}{*}{wood} 
 & AnoVL & {\ul 97.4} & {\ul 99.2} & 96.0 & 95.5 & 77.4 & 57.4 & 51.7 & 34.9 \\
 & AnomalyCLIP & 96.9 & 99.2 & {\ul 96.6} & {\ul 96.5} & \textbf{94.8} & {\ul 61.9} & {\ul 59.1} & {\ul 42.0} \\
 & AdaCLIP & 97.3 & 99.1 & 96.0 & 91.6 & 16.7 & 47.2 & 11.0 & 5.8 \\
 & HeadCLIP & \textbf{98.9} & \textbf{99.6} & \textbf{96.8} & \textbf{97.5} & {\ul 94.7} & \textbf{69.2} & \textbf{64.6} & \textbf{47.8} \\ 
\midrule
\multirow{4}{*}{zipper} 
 & AnoVL & 93.5 & 98.2 & 92.8 & 94.7 & \textbf{82.5} & 24.9 & {\ul 35.1} & {\ul 21.3} \\
 & AnomalyCLIP & {\ul 97.8} & {\ul 99.4} & {\ul 97.1} & 88.9 & 67.5 & 37.4 & \textbf{43.8} & \textbf{28.1} \\
 & AdaCLIP & \textbf{98.7} & \textbf{99.7} & 97.0 & {\ul 95.7} & 2.6 & \textbf{53.4} & 5.3 & 2.7 \\
 & HeadCLIP & 97.5 & 99.3 & \textbf{97.9} & \textbf{95.7} & {\ul 74.4} & {\ul 41.1} & 33.5 & 20.1 \\

\end{longtable}

\renewcommand{\arraystretch}{1.0}

\renewcommand{\arraystretch}{0.5}

\begin{longtable}{@{}>{\scriptsize}c>{\scriptsize}l*{8}{>{\scriptsize\centering\arraybackslash}p{0.8cm}}@{}}
\caption{ZSAD performance comparison on VisA. Best performance shown in \textbf{bold}, second-best shown in {\ul underline}.}
\label{tab:appendix_visa} \\
\toprule
\multirow{2}{*}{\scriptsize Subclass} & \multirow{2}{*}{\scriptsize Model} & \multicolumn{3}{c}{\scriptsize Image-level} & \multicolumn{5}{c}{\scriptsize Pixel-level} \\ 
\cmidrule(lr){3-5} \cmidrule(lr){6-10}
 & & ROC & AP & $\text{F1}_{\text{max}}$ & ROC & PRO & AP & $\text{F1}_{\text{max}}$ & $\text{IoU}_{\text{max}}$ \\ 
\midrule
\endfirsthead

\multicolumn{10}{c}{\tablename\ \thetable{} -- continued from previous page} \\
\toprule
\multirow{2}{*}{\scriptsize Subclass} & \multirow{2}{*}{\scriptsize Model} & \multicolumn{3}{c}{\scriptsize Image-level} & \multicolumn{5}{c}{\scriptsize Pixel-level} \\ 
\cmidrule(lr){3-5} \cmidrule(lr){6-10}
 & & ROC & AP & $\text{F1}_{\text{max}}$ & ROC & PRO & AP & $\text{F1}_{\text{max}}$ & $\text{IoU}_{\text{max}}$ \\ 
\midrule
\endhead

\multicolumn{10}{r}{\scriptsize Continued on next page} \\
\endfoot

\bottomrule
\endlastfoot

\multirow{4}{*}{\scriptsize candle} 
 & AnoVL & {\ul 94.7} & {\ul 95.3} & \textbf{90.7} & 95.1 & 82.9 & 6.6 & 13.9 & 7.5 \\
 & AnomalyCLIP & 67.8 & 71.7 & 68.1 & 98.3 & {\ul 96.2} & {\ul 27.0} & \textbf{38.2} & \textbf{23.6} \\
 & AdaCLIP & \textbf{95.7} & \textbf{96.3} & {\ul 90.4} & \textbf{99.0} & 48.0 & \textbf{32.8} & 33.7 & 20.3 \\
 & HeadCLIP & 84.8 & 87.1 & 78.4 & {\ul 98.8} & \textbf{96.6} & 23.8 & {\ul 37.5} & {\ul 23.1} \\ \midrule
\multirow{4}{*}{\scriptsize capsules} 
 & AnoVL & 76.5 & 84.9 & 81.4 & 84.8 & 46.2 & 2.7 & 8.7 & 4.5 \\
 & AnomalyCLIP & {\ul 83.0} & {\ul 89.6} & {\ul 84.2} & 94.4 & {\ul 87.9} & 16.5 & 24.3 & 13.8 \\
 & AdaCLIP & 80.5 & 88.5 & 80.4 & \textbf{98.3} & 51.2 & \textbf{43.0} & \textbf{49.4} & \textbf{32.8} \\
 & HeadCLIP & \textbf{94.6} & \textbf{97.0} & \textbf{90.6} & {\ul 96.4} & \textbf{90.9} & {\ul 34.4} & {\ul 42.8} & {\ul 27.2} \\ \midrule
\multirow{4}{*}{\scriptsize cashew}  
 & AnoVL & {\ul 89.9} & 95.4 & {\ul 87.8} & 89.4 & 85.7 & 6.9 & 9.7 & 5.1 \\
 & AnomalyCLIP & 81.2 & 91.4 & 82.3 & 89.1 & {\ul 90.6} & 15.8 & {\ul 20.3} & {\ul 11.3} \\
 & AdaCLIP & \textbf{90.3} & \textbf{95.8} & \textbf{88.6} & {\ul 93.8} & 43.1 & {\ul 27.6} & 17.6 & 9.6 \\
 & HeadCLIP & 89.0 & {\ul 95.4} & 86.3 & \textbf{96.7} & \textbf{94.9} & \textbf{35.6} & \textbf{40.2} & \textbf{25.2} \\ \midrule
\multirow{4}{*}{\scriptsize chewinggum} 
 & AnoVL & 95.8 & 98.0 & 91.4 & 96.6 & 74.6 & 45.3 & 50.2 & 33.5 \\
 & AnomalyCLIP & 97.1 & 98.7 & 94.5 & {\ul 99.5} & {\ul 93.9} & {\ul 77.9} & {\ul 72.4} & {\ul 56.7} \\
 & AdaCLIP & {\ul 97.1} & {\ul 98.8} & {\ul 95.4} & \textbf{99.6} & 46.7 & \textbf{84.3} & \textbf{76.7} & \textbf{62.2} \\
 & HeadCLIP & \textbf{97.9} & \textbf{99.2} & \textbf{96.9} & 99.4 & \textbf{94.3} & 61.9 & 65.1 & 48.3 \\ \midrule
\multirow{4}{*}{\scriptsize fryum} 
 & AnoVL & 87.7 & 94.0 & 86.6 & 86.6 & 68.2 & 19.2 & {\ul 29.3} & {\ul 17.2} \\
 & AnomalyCLIP & 88.5 & 94.5 & 86.5 & 92.8 & \textbf{86.2} & 18.3 & 23.8 & 13.5 \\
 & AdaCLIP & {\ul 91.1} & {\ul 96.0} & {\ul 89.4} & {\ul 93.1} & 31.6 & {\ul 21.3} & 12.0 & 6.4 \\
 & HeadCLIP & \textbf{91.7} & \textbf{96.1} & \textbf{90.4} & \textbf{94.6} & {\ul 84.6} & \textbf{27.6} & \textbf{33.4} & \textbf{20.1} \\ \midrule
\multirow{4}{*}{\scriptsize macaroni1} 
 & AnoVL & 68.6 & 71.0 & 70.4 & 85.6 & 55.1 & 0.2 & 1.2 & 0.6 \\
 & AnomalyCLIP & 82.3 & 81.2 & {\ul 78.8} & 98.7 & \textbf{93.7} & {\ul 21.0} & {\ul 30.4} & {\ul 17.9} \\
 & AdaCLIP & {\ul 83.6} & {\ul 84.0} & 78.5 & \textbf{99.3} & 43.5 & \textbf{22.1} & \textbf{30.7} & \textbf{18.1} \\
 & HeadCLIP & \textbf{87.5} & \textbf{87.9} & \textbf{82.0} & {\ul 98.8} & {\ul 93.0} & 11.2 & 21.1 & 11.8 \\ \midrule
\multirow{4}{*}{\scriptsize macaroni2} 
 & AnoVL & 64.9 & 64.1 & {\ul 68.3} & 80.6 & 41.1 & 0.1 & 0.2 & 0.1 \\
 & AnomalyCLIP & {\ul 67.9} & {\ul 65.5} & 68.0 & 98.3 & {\ul 87.6} & 1.9 & 4.7 & 2.4 \\
 & AdaCLIP & 58.9 & 61.9 & 67.3 & \textbf{98.7} & 49.6 & \textbf{9.3} & \textbf{16.7} & \textbf{9.1} \\
 & HeadCLIP & \textbf{71.3} & \textbf{69.1} & \textbf{71.9} & {\ul 98.4} & \textbf{88.3} & {\ul 4.1} & {\ul 8.4} & {\ul 4.4} \\ \midrule
\multirow{4}{*}{\scriptsize pcb1} 
 & AnoVL & 77.9 & 78.1 & 77.7 & \textbf{94.3} & 76.0 & \textbf{13.7} & \textbf{25.9} & \textbf{14.9} \\
 & AnomalyCLIP & 83.6 & 84.5 & 78.3 & 87.2 & {\ul 78.7} & 4.5 & 8.6 & 4.5 \\
 & AdaCLIP & \textbf{85.7} & \textbf{86.9} & \textbf{80.0} & 92.1 & 40.2 & 5.7 & 10.1 & 5.3 \\
 & HeadCLIP & {\ul 85.2} & {\ul 85.7} & {\ul 79.1} & {\ul 92.7} & \textbf{84.6} & {\ul 6.3} & {\ul 10.3} & {\ul 5.4} \\ \midrule
\multirow{4}{*}{\scriptsize pcb2} 
 & AnoVL & 48.0 & 46.5 & 68.1 & 87.6 & 59.6 & 1.3 & 3.4 & 1.7 \\
 & AnomalyCLIP & 63.1 & 64.1 & 68.7 & {\ul 91.1} & {\ul 76.1} & 8.0 & 16.0 & 8.7 \\
 & AdaCLIP & \textbf{75.1} & \textbf{75.8} & \textbf{74.2} & 90.7 & 46.6 & \textbf{18.2} & \textbf{28.9} & \textbf{16.9} \\
 & HeadCLIP & {\ul 70.1} & {\ul 71.3} & {\ul 71.3} & \textbf{92.1} & \textbf{80.7} & {\ul 9.2} & {\ul 16.6} & {\ul 9.0} \\ \midrule
\multirow{4}{*}{\scriptsize pcb3} 
 & AnoVL & 63.6 & 62.4 & 67.4 & 78.2 & 39.1 & 0.6 & 1.3 & 0.6 \\
 & AnomalyCLIP & 67.8 & 73.0 & 67.4 & {\ul 89.7} & {\ul 74.0} & 5.5 & 9.7 & 5.1 \\
 & AdaCLIP & \textbf{78.7} & \textbf{82.7} & \textbf{75.0} & \textbf{90.1} & 42.1 & \textbf{25.1} & \textbf{31.9} & \textbf{19.0} \\
 & HeadCLIP & {\ul 69.3} & {\ul 74.5} & {\ul 67.7} & 89.7 & \textbf{78.3} & {\ul 11.3} & {\ul 19.8} & {\ul 11.0} \\ \midrule
\multirow{4}{*}{\scriptsize pcb4} 
 & AnoVL & 75.0 & 74.0 & 72.8 & 93.0 & {\ul 78.0} & 23.3 & {\ul 31.1} & {\ul 18.4} \\
 & AnomalyCLIP & 94.2 & 94.7 & 87.2 & 92.7 & 77.6 & 17.0 & 23.8 & 13.5 \\
 & AdaCLIP & {\ul 95.9} & {\ul 96.3} & {\ul 90.9} & \textbf{96.0} & 63.8 & \textbf{36.3} & \textbf{40.1} & \textbf{25.1} \\
 & HeadCLIP & \textbf{97.3} & \textbf{97.1} & \textbf{92.8} & {\ul 94.2} & \textbf{84.0} & {\ul 24.1} & 30.0 & 17.6 \\ \midrule
\multirow{4}{*}{\scriptsize pipe fryum} 
 & AnoVL & 79.1 & 89.3 & 82.7 & 86.3 & 89.5 & 5.8 & 12.0 & 6.4 \\
 & AnomalyCLIP & {\ul 94.2} & {\ul 97.4} & {\ul 91.5} & {\ul 96.7} & \textbf{95.9} & {\ul 28.9} & {\ul 36.4} & {\ul 22.3} \\
 & AdaCLIP & 92.5 & 96.4 & 88.8 & 95.8 & 73.3 & 24.0 & 24.9 & 14.2 \\
 & HeadCLIP & \textbf{96.5} & \textbf{98.4} & \textbf{93.6} & \textbf{97.3} & {\ul 95.5} & \textbf{31.2} & \textbf{41.8} & \textbf{26.4} \\

\end{longtable}

\renewcommand{\arraystretch}{1.0}

\renewcommand{\arraystretch}{0.5}

\begin{longtable}{@{}>{\scriptsize}c>{\scriptsize}l*{8}{>{\scriptsize\centering\arraybackslash}p{0.8cm}}@{}}
\caption{ZSAD performance comparison on MPDD. Best performance shown in \textbf{bold}, second-best shown in {\ul underline}.}
\label{tab:appendix_mpdd} \\
\toprule
\multirow{2}{*}{\scriptsize Subclass} & \multirow{2}{*}{\scriptsize Model} & \multicolumn{3}{c}{\scriptsize Image-level} & \multicolumn{5}{c}{\scriptsize Pixel-level} \\ 
\cmidrule(lr){3-5} \cmidrule(lr){6-10}
 & & ROC & AP & $\text{F1}_{\text{max}}$ & ROC & PRO & AP & $\text{F1}_{\text{max}}$ & $\text{IoU}_{\text{max}}$ \\ 
\midrule
\endfirsthead

\multicolumn{10}{c}{\tablename\ \thetable{} -- continued from previous page} \\
\toprule
\multirow{2}{*}{\scriptsize Subclass} & \multirow{2}{*}{\scriptsize Model} & \multicolumn{3}{c}{\scriptsize Image-level} & \multicolumn{5}{c}{\scriptsize Pixel-level} \\ 
\cmidrule(lr){3-5} \cmidrule(lr){6-10}
 & & ROC & AP & $\text{F1}_{\text{max}}$ & ROC & PRO & AP & $\text{F1}_{\text{max}}$ & $\text{IoU}_{\text{max}}$ \\ 
\midrule
\endhead

\multicolumn{10}{r}{\scriptsize Continued on next page} \\
\endfoot

\bottomrule
\endlastfoot

\multirow{4}{*}{\scriptsize bracket black} 
 & AnoVL & 42.8 & 55.6 & 74.6 & 32.4 & 2.4 & 0.1 & 0.2 & 0.1 \\
 & AnomalyCLIP & 49.7 & 58.7 & 75.2 & \textbf{96.4} & \textbf{89.0} & \textbf{8.7} & \textbf{16.4} & \textbf{8.9} \\
 & AdaCLIP & \textbf{72.2} & \textbf{78.0} & \textbf{78.3} & 94.6 & 33.3 & 4.8 & 12.2 & 6.5 \\
 & HeadCLIP & {\ul 61.2} & {\ul 65.4} & {\ul 76.7} & {\ul 95.6} & {\ul 87.1} & {\ul 4.9} & {\ul 13.1} & {\ul 7.0} \\ \midrule
\multirow{4}{*}{\scriptsize bracket brown} 
 & AnoVL & 45.2 & 67.8 & 79.7 & 31.5 & 3.7 & 0.3 & 0.9 & 0.5 \\
 & AnomalyCLIP & \textbf{64.7} & \textbf{80.5} & {\ul 81.0} & 93.8 & \textbf{90.3} & {\ul 4.7} & 9.4 & 4.9 \\
 & AdaCLIP & 52.9 & 71.5 & 80.3 & \textbf{94.3} & 17.3 & \textbf{6.0} & \textbf{11.0} & \textbf{5.8} \\
 & HeadCLIP & {\ul 62.2} & {\ul 75.4} & \textbf{81.3} & {\ul 93.9} & {\ul 84.7} & 4.7 & {\ul 9.4} & {\ul 4.9} \\ \midrule
\multirow{4}{*}{\scriptsize bracket white} 
 & AnoVL & 40.9 & 53.4 & 67.4 & 58.4 & 12.0 & 0.0 & 0.1 & 0.1 \\
 & AnomalyCLIP & 70.7 & \textbf{76.1} & 69.8 & \textbf{99.8} & {\ul 97.7} & \textbf{16.9} & \textbf{26.8} & \textbf{15.5} \\
 & AdaCLIP & \textbf{76.3} & {\ul 75.5} & \textbf{78.9} & 98.1 & 14.3 & 1.3 & 3.5 & 1.8 \\
 & HeadCLIP & {\ul 72.4} & 68.7 & {\ul 76.2} & {\ul 99.6} & \textbf{97.8} & {\ul 16.3} & {\ul 26.6} & {\ul 15.3} \\ \midrule
\multirow{4}{*}{\scriptsize connector} 
 & AnoVL & 77.9 & 62.3 & 64.9 & 78.6 & 38.1 & 1.4 & 3.2 & 1.6 \\
 & AnomalyCLIP & \textbf{86.9} & \textbf{76.8} & \textbf{72.7} & 95.0 & \textbf{83.8} & 14.9 & 20.2 & 11.2 \\
 & AdaCLIP & 74.8 & 63.9 & 62.9 & \textbf{97.2} & 46.4 & \textbf{41.1} & \textbf{42.0} & \textbf{26.6} \\
 & HeadCLIP & {\ul 78.6} & {\ul 70.9} & {\ul 66.7} & {\ul 95.1} & {\ul 83.3} & {\ul 21.9} & {\ul 25.4} & {\ul 14.6} \\ \midrule
\multirow{4}{*}{\scriptsize metal plate} 
 & AnoVL & \textbf{97.2} & \textbf{98.9} & \textbf{96.6} & \textbf{95.5} & {\ul 85.5} & \textbf{69.0} & \textbf{68.5} & \textbf{52.1} \\
 & AnomalyCLIP & 69.3 & 88.7 & 84.5 & 94.6 & \textbf{87.6} & {\ul 63.4} & {\ul 64.5} & {\ul 47.6} \\
 & AdaCLIP & 84.0 & 94.3 & 87.6 & {\ul 95.0} & 38.4 & 62.2 & 36.2 & 22.1 \\
 & HeadCLIP & {\ul 94.6} & {\ul 97.9} & {\ul 93.8} & 92.2 & 83.2 & 60.4 & 56.6 & 39.5 \\ \midrule
\multirow{4}{*}{\scriptsize tubes} 
 & AnoVL & 86.3 & 93.7 & 86.8 & 93.2 & 75.5 & 11.7 & 17.7 & 9.7 \\
 & AnomalyCLIP & 95.4 & 97.9 & {\ul 92.8} & 97.1 & {\ul 90.1} & 32.0 & 33.4 & 20.1 \\
 & AdaCLIP & \textbf{99.6} & \textbf{99.8} & \textbf{97.9} & \textbf{99.1} & 70.4 & \textbf{71.2} & \textbf{66.8} & \textbf{50.2} \\
 & HeadCLIP & {\ul 95.6} & {\ul 98.0} & 92.1 & {\ul 97.7} & \textbf{91.3} & {\ul 47.4} & {\ul 47.1} & {\ul 30.8} \\

\end{longtable}

\renewcommand{\arraystretch}{1.0}

\renewcommand{\arraystretch}{0.5}

\begin{longtable}{@{}>{\scriptsize}c>{\scriptsize}l*{8}{>{\scriptsize\centering\arraybackslash}p{0.8cm}}@{}}
\caption{ZSAD performance comparison on BTAD. Best performance shown in \textbf{bold}, second-best shown in {\ul underline}.}
\label{tab:appendix_btad} \\
\toprule
\multirow{2}{*}{\scriptsize Subclass} & \multirow{2}{*}{\scriptsize Model} & \multicolumn{3}{c}{\scriptsize Image-level} & \multicolumn{5}{c}{\scriptsize Pixel-level} \\ 
\cmidrule(lr){3-5} \cmidrule(lr){6-10}
 & & ROC & AP & $\text{F1}_{\text{max}}$ & ROC & PRO & AP & $\text{F1}_{\text{max}}$ & $\text{IoU}_{\text{max}}$ \\ 
\midrule
\endfirsthead

\multicolumn{10}{c}{\tablename\ \thetable{} -- continued from previous page} \\
\toprule
\multirow{2}{*}{\scriptsize Subclass} & \multirow{2}{*}{\scriptsize Model} & \multicolumn{3}{c}{\scriptsize Image-level} & \multicolumn{5}{c}{\scriptsize Pixel-level} \\ 
\cmidrule(lr){3-5} \cmidrule(lr){6-10}
 & & ROC & AP & $\text{F1}_{\text{max}}$ & ROC & PRO & AP & $\text{F1}_{\text{max}}$ & $\text{IoU}_{\text{max}}$ \\ 
\midrule
\endhead

\multicolumn{10}{r}{\scriptsize Continued on next page} \\
\endfoot

\bottomrule
\endlastfoot

\multirow{4}{*}{\scriptsize 01} 
 & AnoVL & 89.5 & 96.3 & 89.9 & {\ul 94.0} & {\ul 58.7} & {\ul 41.8} & {\ul 47.4} & {\ul 31.1} \\
 & AnomalyCLIP & {\ul 96.3} & {\ul 98.6} & {\ul 93.6} & 84.3 & 47.7 & 21.9 & 25.6 & 14.7 \\
 & AdaCLIP & 52.8 & 80.8 & 82.4 & 83.9 & 5.9 & 39.9 & 16.5 & 9.0 \\
 & HeadCLIP & \textbf{96.7} & \textbf{98.9} & \textbf{96.8} & \textbf{94.2} & \textbf{79.1} & \textbf{54.4} & \textbf{58.1} & \textbf{40.9} \\ \midrule
\multirow{4}{*}{\scriptsize 02} 
 & AnoVL & 59.6 & 92.1 & 93.0 & 79.2 & 26.5 & 24.3 & 31.7 & 18.8 \\
 & AnomalyCLIP & 81.2 & 96.9 & {\ul 93.2} & 94.8 & {\ul 69.8} & 65.1 & {\ul 62.3} & {\ul 45.3} \\
 & AdaCLIP & {\ul 84.4} & {\ul 97.4} & \textbf{94.3} & \textbf{96.6} & 43.0 & \textbf{71.4} & 35.7 & 21.7 \\
 & HeadCLIP & \textbf{88.0} & \textbf{98.1} & {\ul 93.2} & {\ul 95.9} & \textbf{72.8} & {\ul 69.8} & \textbf{66.0} & \textbf{49.3} \\ \midrule
\multirow{4}{*}{\scriptsize 03} 
 & AnoVL & 66.8 & 16.3 & 25.0 & 86.1 & 69.1 & 3.8 & 9.3 & 4.9 \\
 & AnomalyCLIP & 83.8 & 51.4 & 50.7 & 93.5 & {\ul 86.1} & 6.5 & 14.1 & 7.6 \\
 & AdaCLIP & {\ul 98.0} & {\ul 91.4} & {\ul 87.2} & {\ul 95.4} & 28.4 & \textbf{28.0} & \textbf{29.6} & \textbf{17.4} \\
 & HeadCLIP & \textbf{98.9} & \textbf{94.6} & \textbf{91.6} & \textbf{96.6} & \textbf{92.2} & {\ul 18.8} & {\ul 28.5} & {\ul 16.6} \\

\end{longtable}

\renewcommand{\arraystretch}{1.0}

\renewcommand{\arraystretch}{0.5}

\begin{longtable}{@{}>{\scriptsize}c>{\scriptsize}l*{8}{>{\scriptsize\centering\arraybackslash}p{0.8cm}}@{}}
\caption{ZSAD performance comparison on DAGM. Best performance shown in \textbf{bold}, second-best shown in {\ul underline}.}
\label{tab:appendix_dagm} \\
\toprule
\multirow{2}{*}{\scriptsize Subclass} & \multirow{2}{*}{\scriptsize Model} & \multicolumn{3}{c}{\scriptsize Image-level} & \multicolumn{5}{c}{\scriptsize Pixel-level} \\ 
\cmidrule(lr){3-5} \cmidrule(lr){6-10}
 & & ROC & AP & $\text{F1}_{\text{max}}$ & ROC & PRO & AP & $\text{F1}_{\text{max}}$ & $\text{IoU}_{\text{max}}$ \\ 
\midrule
\endfirsthead

\multicolumn{10}{c}{\tablename\ \thetable{} -- continued from previous page} \\
\toprule
\multirow{2}{*}{\scriptsize Subclass} & \multirow{2}{*}{\scriptsize Model} & \multicolumn{3}{c}{\scriptsize Image-level} & \multicolumn{5}{c}{\scriptsize Pixel-level} \\ 
\cmidrule(lr){3-5} \cmidrule(lr){6-10}
 & & ROC & AP & $\text{F1}_{\text{max}}$ & ROC & PRO & AP & $\text{F1}_{\text{max}}$ & $\text{IoU}_{\text{max}}$ \\ 
\midrule
\endhead

\multicolumn{10}{r}{\scriptsize Continued on next page} \\
\endfoot

\bottomrule
\endlastfoot

\multirow{4}{*}{\scriptsize Class1} 
 & AnoVL & 60.8 & 20.1 & 29.9 & 66.7 & 27.4 & 1.1 & 3.3 & 1.7 \\
 & AnomalyCLIP & 84.4 & 47.7 & 53.8 & 86.1 & {\ul 73.9} & 36.0 & {\ul 41.3} & {\ul 26.0} \\
 & AdaCLIP & {\ul 94.9} & {\ul 77.7} & {\ul 71.5} & {\ul 86.9} & 24.2 & {\ul 47.0} & 21.1 & 11.8 \\
 & HeadCLIP & \textbf{97.6} & \textbf{89.8} & \textbf{82.1} & \textbf{89.6} & \textbf{79.0} & \textbf{53.0} & \textbf{55.9} & \textbf{38.8} \\
\midrule
\multirow{4}{*}{\scriptsize Class2} 
 & AnoVL & 99.5 & 98.2 & 95.7 & 96.5 & 89.6 & 15.9 & 27.4 & 15.9 \\
 & AnomalyCLIP & {\ul 100.0} & {\ul 99.9} & {\ul 99.4} & {\ul 99.2} & {\ul 98.8} & \textbf{72.1} & \textbf{67.6} & \textbf{51.1} \\
 & AdaCLIP & {\ul 100.0} & 99.9 & 98.8 & 97.6 & 31.0 & 66.6 & 27.8 & 16.2 \\
 & HeadCLIP & \textbf{100.0} & \textbf{100.0} & \textbf{100.0} & \textbf{99.6} & \textbf{98.8} & {\ul 71.4} & {\ul 67.5} & {\ul 51.0} \\
\midrule
\multirow{4}{*}{\scriptsize Class3} 
 & AnoVL & 98.9 & 96.1 & 93.3 & 93.8 & 80.4 & 8.9 & 16.9 & 9.2 \\
 & AnomalyCLIP & 99.8 & 99.2 & 96.4 & 91.9 & {\ul 88.5} & 61.3 & {\ul 64.8} & {\ul 48.0} \\
 & AdaCLIP & {\ul 100.0} & {\ul 100.0} & {\ul 98.8} & {\ul 94.9} & 25.9 & {\ul 69.7} & 27.8 & 16.2 \\
 & HeadCLIP & \textbf{100.0} & \textbf{100.0} & \textbf{100.0} & \textbf{95.7} & \textbf{93.5} & \textbf{71.0} & \textbf{70.3} & \textbf{54.2} \\
\midrule
\multirow{4}{*}{\scriptsize Class4} 
 & AnoVL & 94.1 & 83.9 & 76.1 & 86.3 & {\ul 70.0} & \textbf{8.8} & \textbf{16.8} & \textbf{9.2} \\
 & AnomalyCLIP & {\ul 97.1} & {\ul 89.6} & {\ul 83.7} & 85.7 & 65.9 & 4.0 & 12.0 & 6.4 \\
 & AdaCLIP & 91.9 & 73.3 & 67.1 & {\ul 89.3} & 0.5 & 3.2 & 0.8 & 0.4 \\
 & HeadCLIP & \textbf{98.0} & \textbf{92.5} & \textbf{88.4} & \textbf{91.1} & \textbf{77.0} & {\ul 6.8} & {\ul 16.5} & {\ul 9.0} \\
\midrule
\multirow{4}{*}{\scriptsize Class5} 
 & AnoVL & {\ul 95.4} & 88.0 & {\ul 82.7} & 91.3 & 72.1 & 14.2 & 20.1 & 11.2 \\
 & AnomalyCLIP & \textbf{100.0} & \textbf{100.0} & \textbf{100.0} & 95.5 & {\ul 88.2} & 61.6 & {\ul 63.0} & {\ul 46.0} \\
 & AdaCLIP & \textbf{100.0} & \textbf{100.0} & \textbf{100.0} & {\ul 98.3} & 34.8 & \textbf{80.8} & 39.2 & 24.4 \\
 & HeadCLIP & \textbf{100.0} & \textbf{100.0} & \textbf{100.0} & \textbf{98.9} & \textbf{96.4} & {\ul 75.8} & \textbf{73.4} & \textbf{58.0} \\
\midrule
\multirow{4}{*}{\scriptsize Class6} 
 & AnoVL & {\ul 99.8} & {\ul 98.7} & {\ul 94.2} & 95.2 & 85.0 & 42.5 & 46.5 & 30.3 \\
 & AnomalyCLIP & \textbf{100.0} & \textbf{100.0} & \textbf{100.0} & 96.4 & {\ul 90.8} & 75.5 & {\ul 74.4} & {\ul 59.3} \\
 & AdaCLIP & \textbf{100.0} & \textbf{100.0} & \textbf{100.0} & {\ul 98.7} & 41.5 & \textbf{85.5} & 49.5 & 32.9 \\
 & HeadCLIP & \textbf{100.0} & \textbf{100.0} & \textbf{100.0} & \textbf{99.4} & \textbf{96.9} & {\ul 85.1} & \textbf{78.8} & \textbf{65.1} \\
\midrule
\multirow{4}{*}{\scriptsize Class7} 
 & AnoVL & 85.6 & 56.2 & 55.3 & 75.6 & 47.7 & 4.9 & 10.2 & 5.4 \\
 & AnomalyCLIP & 99.9 & 99.5 & 96.9 & 89.7 & {\ul 83.9} & 59.1 & {\ul 63.9} & {\ul 47.0} \\
 & AdaCLIP & \textbf{100.0} & \textbf{100.0} & \textbf{100.0} & \textbf{94.3} & 51.5 & \textbf{75.8} & 56.2 & 39.1 \\
 & HeadCLIP & {\ul 100.0} & {\ul 99.9} & {\ul 98.7} & {\ul 90.8} & \textbf{87.2} & {\ul 64.3} & \textbf{66.8} & \textbf{50.1} \\
\midrule
\multirow{4}{*}{\scriptsize Class8} 
 & AnoVL & 72.7 & 38.3 & 38.7 & 73.2 & 33.9 & 0.3 & 1.6 & 0.8 \\
 & AnomalyCLIP & 94.9 & 87.9 & 82.7 & 95.6 & {\ul 92.3} & 52.1 & {\ul 56.1} & {\ul 39.0} \\
 & AdaCLIP & \textbf{99.7} & \textbf{98.2} & \textbf{96.3} & {\ul 97.3} & 14.4 & \textbf{64.0} & 18.6 & 10.3 \\
 & HeadCLIP & {\ul 97.5} & {\ul 92.8} & {\ul 87.5} & \textbf{98.4} & \textbf{98.0} & {\ul 62.7} & \textbf{62.8} & \textbf{45.8} \\
\midrule
\multirow{4}{*}{\scriptsize Class9} 
 & AnoVL & 94.3 & 77.5 & 69.8 & 96.1 & 87.4 & 2.6 & 7.5 & 3.9 \\
 & AnomalyCLIP & 98.0 & 93.5 & 86.8 & 97.6 & {\ul 93.2} & 42.4 & {\ul 46.1} & {\ul 29.9} \\
 & AdaCLIP & \textbf{99.8} & \textbf{99.0} & \textbf{96.3} & {\ul 97.8} & 43.0 & \textbf{66.8} & 27.1 & 15.7 \\
 & HeadCLIP & {\ul 99.1} & {\ul 94.0} & {\ul 88.7} & \textbf{99.7} & \textbf{98.8} & {\ul 50.0} & \textbf{50.6} & \textbf{33.8} \\
\midrule
\multirow{4}{*}{\scriptsize Class10} 
 & AnoVL & 98.0 & 93.5 & 90.7 & {\ul 98.7} & 96.1 & 19.9 & 28.7 & 16.8 \\
 & AnomalyCLIP & 99.9 & 99.3 & {\ul 97.3} & 98.2 & {\ul 96.7} & {\ul 66.8} & {\ul 64.8} & {\ul 47.9} \\
 & AdaCLIP & {\ul 99.9} & {\ul 99.5} & 96.7 & 97.9 & 18.8 & 56.5 & 11.0 & 5.8 \\
 & HeadCLIP & \textbf{100.0} & \textbf{100.0} & \textbf{100.0} & \textbf{99.2} & \textbf{98.4} & \textbf{70.3} & \textbf{67.8} & \textbf{51.3} \\

\end{longtable}

\renewcommand{\arraystretch}{1.0}

\renewcommand{\arraystretch}{0.5}

\begin{longtable}{@{}>{\scriptsize}c>{\scriptsize}l*{8}{>{\scriptsize\centering\arraybackslash}p{0.8cm}}@{}}
\caption{ZSAD performance comparison on DTD-Synthetic. Best performance shown in \textbf{bold}, second-best shown in {\ul underline}.}
\label{tab:appendix_dtd} \\
\toprule
\multirow{2}{*}{\scriptsize Subclass} & \multirow{2}{*}{\scriptsize Model} & \multicolumn{3}{c}{\scriptsize Image-level} & \multicolumn{5}{c}{\scriptsize Pixel-level} \\ 
\cmidrule(lr){3-5} \cmidrule(lr){6-10}
 & & ROC & AP & $\text{F1}_{\text{max}}$ & ROC & PRO & AP & $\text{F1}_{\text{max}}$ & $\text{IoU}_{\text{max}}$ \\ 
\midrule
\endfirsthead

\multicolumn{10}{c}{\tablename\ \thetable{} -- continued from previous page} \\
\toprule
\multirow{2}{*}{\scriptsize Subclass} & \multirow{2}{*}{\scriptsize Model} & \multicolumn{3}{c}{\scriptsize Image-level} & \multicolumn{5}{c}{\scriptsize Pixel-level} \\ 
\cmidrule(lr){3-5} \cmidrule(lr){6-10}
 & & ROC & AP & $\text{F1}_{\text{max}}$ & ROC & PRO & AP & $\text{F1}_{\text{max}}$ & $\text{IoU}_{\text{max}}$ \\ 
\midrule
\endhead

\multicolumn{10}{r}{\scriptsize Continued on next page} \\
\endfoot

\bottomrule
\endlastfoot

\multirow{4}{*}{\scriptsize Blotchy 099} 
 & AnoVL & 98.3 & 99.6 & 98.1 & 98.4 & 93.5 & 50.1 & 48.1 & 31.7 \\
 & AnomalyCLIP & 97.9 & 99.4 & 98.1 & {\ul 99.6} & \textbf{97.6} & 81.8 & {\ul 76.0} & {\ul 61.3} \\
 & AdaCLIP & \textbf{100.0} & \textbf{100.0} & \textbf{100.0} & 99.6 & 57.7 & {\ul 82.1} & 70.2 & 54.1 \\
 & HeadCLIP & {\ul 99.8} & {\ul 99.9} & {\ul 99.4} & \textbf{99.7} & {\ul 96.8} & \textbf{83.8} & \textbf{76.2} & \textbf{61.5} \\
\midrule
\multirow{4}{*}{\scriptsize Fibrous 183} 
 & AnoVL & 96.8 & 99.2 & 95.1 & 97.9 & 92.1 & 38.9 & 46.4 & 30.2 \\
 & AnomalyCLIP & 98.2 & 99.6 & 97.4 & 99.5 & {\ul 98.7} & {\ul 81.0} & {\ul 75.3} & {\ul 60.4} \\
 & AdaCLIP & {\ul 99.0} & {\ul 99.8} & {\ul 98.1} & {\ul 99.5} & 85.3 & 80.0 & 72.8 & 57.3 \\
 & HeadCLIP & \textbf{99.6} & \textbf{99.9} & \textbf{98.1} & \textbf{99.7} & \textbf{98.7} & \textbf{84.6} & \textbf{77.5} & \textbf{63.2} \\
\midrule
\multirow{4}{*}{\scriptsize Marbled 078} 
 & AnoVL & 97.4 & 99.4 & 96.8 & 98.3 & 93.7 & 46.8 & 50.0 & 33.3 \\
 & AnomalyCLIP & 98.6 & 99.7 & 97.4 & 99.3 & {\ul 98.0} & {\ul 77.7} & \textbf{70.6} & \textbf{54.5} \\
 & AdaCLIP & {\ul 98.9} & {\ul 99.8} & {\ul 98.1} & \textbf{99.6} & 67.9 & \textbf{84.5} & {\ul 69.7} & {\ul 53.5} \\
 & HeadCLIP & \textbf{99.9} & \textbf{100.0} & \textbf{99.4} & {\ul 99.5} & \textbf{98.2} & 74.8 & 69.0 & 52.7 \\
\midrule
\multirow{4}{*}{\scriptsize Matted 069} 
 & AnoVL & \textbf{95.4} & {\ul 98.9} & \textbf{94.9} & 96.7 & 77.3 & 25.2 & 31.8 & 18.9 \\
 & AnomalyCLIP & 80.9 & 94.7 & 90.6 & 99.3 & {\ul 84.3} & 74.4 & {\ul 68.6} & {\ul 52.2} \\
 & AdaCLIP & {\ul 95.1} & \textbf{98.9} & {\ul 94.7} & {\ul 99.3} & 70.4 & {\ul 76.7} & 68.2 & 51.7 \\
 & HeadCLIP & 94.6 & 98.7 & 93.3 & \textbf{99.7} & \textbf{87.5} & \textbf{79.6} & \textbf{72.2} & \textbf{56.5} \\
\midrule
\multirow{4}{*}{\scriptsize Mesh 114} 
 & AnoVL & 79.9 & 91.8 & 83.7 & 88.8 & 64.3 & 23.6 & 31.0 & 18.3 \\
 & AnomalyCLIP & 80.2 & 92.0 & 83.3 & \textbf{96.7} & \textbf{87.9} & 60.5 & {\ul 60.1} & {\ul 42.9} \\
 & AdaCLIP & {\ul 87.9} & {\ul 95.2} & \textbf{87.6} & 95.1 & 36.4 & \textbf{67.5} & 52.7 & 35.8 \\
 & HeadCLIP & \textbf{88.1} & \textbf{95.3} & {\ul 86.4} & {\ul 96.1} & {\ul 82.1} & {\ul 61.6} & \textbf{61.1} & \textbf{44.0} \\
\midrule
\multirow{4}{*}{\scriptsize Perforated 037} 
 & AnoVL & \textbf{99.1} & \textbf{99.8} & \textbf{97.4} & \textbf{98.9} & \textbf{96.1} & 46.2 & 47.6 & 31.2 \\
 & AnomalyCLIP & 88.6 & 96.7 & 93.4 & {\ul 96.9} & 91.4 & {\ul 66.8} & \textbf{65.8} & \textbf{49.0} \\
 & AdaCLIP & {\ul 93.5} & {\ul 98.5} & 93.5 & 96.1 & 22.0 & 66.7 & 31.2 & 18.5 \\
 & HeadCLIP & 92.9 & 98.2 & {\ul 93.8} & 96.0 & {\ul 92.5} & \textbf{67.3} & {\ul 64.3} & {\ul 47.4} \\
\midrule
\multirow{4}{*}{\scriptsize Stratified 154} 
 & AnoVL & {\ul 98.8} & {\ul 99.7} & 97.5 & 99.3 & {\ul 94.7} & 68.1 & 61.6 & 44.5 \\
 & AnomalyCLIP & 98.2 & 99.6 & 96.8 & {\ul 99.5} & \textbf{98.9} & \textbf{81.3} & \textbf{73.7} & \textbf{58.4} \\
 & AdaCLIP & \textbf{99.6} & \textbf{99.9} & \textbf{98.8} & 99.1 & 40.4 & 77.2 & 27.2 & 15.8 \\
 & HeadCLIP & 98.7 & 99.7 & {\ul 98.1} & \textbf{99.7} & 92.0 & {\ul 78.4} & {\ul 72.3} & {\ul 56.6} \\
\midrule
\multirow{4}{*}{\scriptsize Woven 001} 
 & AnoVL & {\ul 94.8} & {\ul 98.0} & {\ul 92.8} & 96.1 & 79.7 & 27.9 & 34.5 & 20.8 \\
 & AnomalyCLIP & \textbf{100.0} & \textbf{100.0} & \textbf{100.0} & 99.7 & \textbf{99.1} & {\ul 77.8} & {\ul 74.8} & {\ul 59.8} \\
 & AdaCLIP & \textbf{100.0} & \textbf{100.0} & \textbf{100.0} & \textbf{99.8} & 75.0 & \textbf{85.0} & \textbf{76.1} & \textbf{61.5} \\
 & HeadCLIP & \textbf{100.0} & \textbf{100.0} & \textbf{100.0} & {\ul 99.8} & {\ul 98.2} & 74.7 & 69.6 & 53.4 \\
\midrule
\multirow{4}{*}{\scriptsize Woven 068} 
 & AnoVL & 89.0 & 93.7 & 85.5 & 97.8 & 90.9 & 27.6 & 36.2 & 22.1 \\
 & AnomalyCLIP & {\ul 96.9} & {\ul 98.4} & {\ul 94.9} & {\ul 98.9} & {\ul 96.0} & {\ul 62.1} & {\ul 58.8} & {\ul 41.6} \\
 & AdaCLIP & \textbf{97.0} & \textbf{98.5} & \textbf{96.2} & 98.5 & 69.8 & \textbf{74.5} & \textbf{67.2} & \textbf{50.6} \\
 & HeadCLIP & 96.5 & 98.1 & 92.7 & \textbf{99.2} & \textbf{96.8} & 59.6 & 57.4 & 40.3 \\
\midrule
\multirow{4}{*}{\scriptsize Woven 104} 
 & AnoVL & 96.0 & 99.0 & 95.2 & 98.0 & 92.7 & 45.3 & 43.9 & 28.1 \\
 & AnomalyCLIP & {\ul 97.7} & {\ul 99.5} & {\ul 97.4} & 96.5 & {\ul 93.3} & 68.4 & {\ul 65.3} & {\ul 48.4} \\
 & AdaCLIP & 97.6 & 99.4 & 96.2 & \textbf{98.3} & 49.9 & \textbf{74.5} & 55.1 & 38.0 \\
 & HeadCLIP & \textbf{99.5} & \textbf{99.9} & \textbf{98.7} & {\ul 98.2} & \textbf{95.5} & {\ul 73.6} & \textbf{66.6} & \textbf{50.0} \\
\midrule
\multirow{4}{*}{\scriptsize Woven 125} 
 & AnoVL & 96.5 & 99.2 & 96.1 & 97.3 & 91.8 & 31.0 & 39.2 & 24.4 \\
 & AnomalyCLIP & {\ul 99.7} & {\ul 99.9} & {\ul 98.7} & 99.6 & \textbf{98.5} & {\ul 80.7} & \textbf{72.9} & \textbf{57.4} \\
 & AdaCLIP & \textbf{100.0} & \textbf{100.0} & \textbf{100.0} & \textbf{99.8} & 66.3 & \textbf{88.5} & 71.1 & 55.1 \\
 & HeadCLIP & \textbf{100.0} & \textbf{100.0} & \textbf{100.0} & {\ul 99.6} & {\ul 98.1} & 77.3 & {\ul 71.7} & {\ul 55.9} \\
\midrule
\multirow{4}{*}{\scriptsize Woven 127} 
 & AnoVL & 83.1 & 84.4 & 79.3 & 89.7 & 77.0 & 12.8 & 21.1 & 11.8 \\
 & AnomalyCLIP & 79.2 & 83.1 & 75.5 & 94.3 & {\ul 87.7} & 48.6 & {\ul 52.4} & {\ul 35.5} \\
 & AdaCLIP & {\ul 88.3} & {\ul 89.2} & {\ul 84.5} & {\ul 94.6} & 55.5 & \textbf{61.8} & 51.4 & 34.6 \\
 & HeadCLIP & \textbf{93.8} & \textbf{95.4} & \textbf{88.0} & \textbf{95.6} & \textbf{92.0} & {\ul 56.0} & \textbf{55.9} & \textbf{38.8} \\

\end{longtable}

\renewcommand{\arraystretch}{1.0}

\subsection{TextSpan Analysis of Learned Head Weights}
\label{appendix:textspan_analysis}

To interpret the semantic specialization of attention heads selected by the learnable head weights, the TextSpan methodology~\citep{textspan} is employed. TextSpan identifies the visual concepts that each attention head responds to by finding text descriptions that maximally activate each head. This analysis reveals why certain heads are amplified (weight $> 1$) or suppressed (weight $\leq 1$) for industrial anomaly detection.

Table~\ref{tab:suppressed_heads} presents the layer-wise analysis of suppressed attention heads. These heads predominantly encode task-irrelevant features including geographic locations (e.g., washington, connecticut, korea), web/commerce terminology (e.g., social, wildlife, bluetooth), and brand names (e.g., motorola, nokia, ericsson). Notably, heads containing the keyword ``inspection'' are correctly suppressed because their contexts are contaminated with irrelevant noise terms, indicating that surface-level keyword matching alone is insufficient for anomaly detection.

Table~\ref{tab:amplified_heads} shows the amplified attention heads across transformer layers. A clear hierarchical pattern emerges: early layers (5--7) capture basic structural and scene-type features; middle layers (8--14) progressively encode color spectrum, material properties, and object categories; late layers (15--19) specialize in fine-grained textures, surface patterns, and spatial structures critical for anomaly localization. This progression validates that effective industrial anomaly detection relies on multi-scale texture representations rather than high-level semantic features.

\renewcommand{\arraystretch}{0.5}
\setlength{\tabcolsep}{3pt}

\begin{longtable}{@{}>{\scriptsize}p{1.2cm}>{\scriptsize\raggedright}p{9cm}>{\scriptsize\raggedright\arraybackslash}p{4cm}@{}}
\caption{Layer-wise analysis of suppressed attention heads (w$\leq$1) based on TextSpan analysis.}
\label{tab:suppressed_heads} \\
\toprule
\textbf{Layer} & \textbf{Key Descriptors} & \textbf{Interpreted Irrelevant Features} \\ 
\midrule
\endfirsthead

\multicolumn{3}{c}{\tablename\ \thetable{} -- continued from previous page} \\
\toprule
\textbf{Layer} & \textbf{Key Descriptors} & \textbf{Interpreted Irrelevant Features} \\ 
\midrule
\endhead

\multicolumn{3}{r}{\scriptsize Continued on next page} \\
\endfoot

\bottomrule
\endlastfoot

Layer 5 & hp, iraq, notebook, social, tree, creek, enterprises, lingerie, rings, bluetooth & Generic web terms \& product names \\
\midrule
Layer 6 & washington, sql, cambridge, keys, uk, lcd, google, size, buildings, hawaii, newspaper, iran, sun, arkansas, nokia, netherlands, wildlife, social, enable, connecticut & Geographic locations, web/tech terms, brand names \\
\midrule
Layer 7 & pink, mac, maryland, ford, spa, motorola, wildlife, meetings, indiana, scott, workshop, portal, iran, buses, diseases, phones, bondage, catalog, bluetooth & Geographic, brand names, abstract terms \\
\midrule
Layer 8 & minimum, vehicles, profile, ericsson, painting, lcd, enable, dallas, psp, tony, rent, lingerie, cambridge, washington, desk, cell, flowers, laptop, dakota, netherlands, tripadvisor, golf & Mixed: some color/size but dominated by geographic \& commerce \\
\midrule
Layer 9 & displayed, reservations, webmaster, wildlife, indiana, ram, cable, sydney, social, yellow, ericsson, avg, red, maine, aol, ink, trees, citizens, phentermine, nude, interviews, matches, programs, psp, postal, census, rentals, outdoors & Geographic, commerce, abstract functional terms \\
\midrule
Layer 10 & michigan, plain, enable, wildlife, input, controls, lingerie, platform, revision, contacts, listings, davis, lcd, iran, elementary, trademarks, enterprise, photography, tennis, draft, tripadvisor, metal, msn, cnet, planning, boston, utilities, navy, census, miles, permanent, aol, compatible, missouri, bedroom & Geographic, commerce, travel-related terms \\
\midrule
Layer 11 & paris, specialist, psp, bluetooth, social, flights, creek, trademarks, lingerie, serial, tags, pool, guess, lcd, photography, connecticut, diego, naked, belgium, costa, owner, monitor, austin, gnu, diseases, reservations, visa, painting, hosting, washington, attractions, bondage, guitar, administration, grants, missouri, device, visitors & Geographic, commerce, general web content \\
\midrule
Layer 12 & interior, draft, wildlife, photography, guess, red, provider, yellow, creek, clearance, silver, maintained, safari, lane, washington, avg, social, rom, green, utc, black, blue, boxes, circuit, epinions, certificates, gnu, lingerie, glass, pink, leather, cvs, vhs, connecticut, iran, dates, browser, bedroom, steps, motorola, cambridge, category, atom, pdt, painting & Geographic, commerce, mixed color (redundant) \\
\midrule
Layer 13 & rom, blue, magazine, advisory, netherlands, white, ringtones, aa, red, avg, category, wood, green, epinions, yellow, washington, index, efforts, windows, draft, creek, application, film, ordering, silver, cable, pink, availability, cvs, baseball, safari, elementary, publications, black, overall, kentucky, plate, profiles, defense, affiliates, branch, lcd, dakota, blogs, trademarks & Color (redundant with L16H2), geographic, commerce \\
\midrule
Layer 14 & social, details, washington, dvd, overall, mirror, prints, installed, epinions, cable, elementary, profile, ringtones, merchant, wildlife, chair, creek, hat, rain, outdoors, hall, film, avg, specialist, photography, revision, yellow, card, psp, motion, middle, posters, width, dsl, pubmed, profiles, vendor, rom, lingerie, column, lcd, tables, map, preview, bond, eur, protected, bears, editor, zealand, patch, letters, bondage, sun, transaction, clubs, oregon, flash, monitor, banks, publications, glass, inspection, gnu, overall, role, kong, cambridge & Geographic, commerce, web content, abstract terms \\
\midrule
Layer 15 & social, affairs, portal, red, enable, kong, belgium, avg, defense, cable, src, arms, dakota, wildlife, utilities, dealers, size, pdt, ericsson, dsl, servers, gnu, painting, williams, flights, felt, leather, ski, panasonic, shirts, newspaper, film, lesbians, citizens, plate, cambridge, sweden, miles, york, installed, spread, bluetooth, circuit, contained, iowa, appliances, initial, room, background, bought, costa, editorial, tables, flights, carrier, discussions, administrator, msn, rear, mirror, chair, orange, printable, howard, nude, magazines, logo, rings, interior, departments, processor & Geographic, brand names, commerce, general attributes \\
\midrule
Layer 16 & green, orange, silver, blue, yellow, pink, black, brown, red, white (color-redundant), net, quote, boards, bath, application, hat, specialist, directory, social, iran, connecticut, card, trademarks, wildlife, washington, editorial, servers, palm, cable, guitar, wales, anderson, grants, epinions, maine, logo, lighting, jones, floor, portal, painting, washington, pda, cvs, clips, outdoors, portal, hawaii, ordering, michigan, chair, bluetooth, poster, laptops, singapore, restaurant, oklahoma, sun, night, netherlands, bedroom, cnet, flights & Geographic, web/commerce, redundant color encoding \\
\midrule
Layer 17 & red, cheats, cent, missouri, wildlife, orange, ram, scotland, pda, draft, rear, eyes, sizes, profiles, washington, mg, rom, overall, patients, forces, desk, sc, enable, screen, label, binding, wildlife, cvs, ann, psp, creek, defense, seattle, contacts, ratio, codes, size, posters, cable, black, details, presentation, compatible, profile, interior, posters, driver, hat, pack, defense, cvs, paris, motorola, feet, ocean, painting, laser, newspaper, television, client, film, belgium, poster & Geographic, web/commerce, abstract functional terms \\
\midrule
Layer 18 & mirror, hands, hills, rom, connecticut, newsletter, flights, czech, gmt, cable, guitar, rain, sun, evening, hall, outdoors, korea, recovery, maine, sky, inspection*, chair, shirt, cell, discussions, pregnancy, running, rear, sign, hat, driving, social, lcd, creek, overall, kong, michigan, cable, inspection*, instruments, henry, flights, containing, floor, wildlife, front, partners, discussions & Geographic, temporal/weather, noisy `inspection' context \\
\midrule
Layer 19 & korea, photography, citizens, connecticut, pdt, cable, social, scotland, spa, wildlife, fixed, bears, draft, size, participants, available, publication, screen, pennsylvania, floor, rear, indicated, utc, vendor, installed, enterprise, wildlife, massachusetts, column, elementary, results, euro, psp, visit, line, forums, dvd, newsletters, navy, speakers, philadelphia, chocolate, driving, bluetooth, supplies, playstation, bedroom, cable, audience, mesh, specialist, book, window, steps, missouri, kitchen, profile, objectives, motion, unit, inspection*, hand, plants, waiting, participants, motorola, bedroom, servers, visit, film, korea, uk, selling, posters, bookmark, box, appliances, sign, card, hardcover, screen, hat, utc & Geographic, consumer electronics, web content, noisy `inspection' \\

\end{longtable}

\setlength{\tabcolsep}{6pt}
\renewcommand{\arraystretch}{1.0}

\renewcommand{\arraystretch}{0.5}
\setlength{\tabcolsep}{3pt}

\begin{longtable}{@{}>{\scriptsize}p{1.2cm}>{\scriptsize\raggedright}p{7cm}>{\scriptsize\raggedright}p{3cm}>{\scriptsize\raggedright\arraybackslash}p{3.5cm}@{}}
\caption{Amplified attention heads (weight $> 1$) analysis by layer. Key descriptors extracted via TextSpan methodology.}
\label{tab:amplified_heads} \\
\toprule
\textbf{Layer} & \textbf{Key Descriptors} & \textbf{Interpreted Specialization} & \textbf{Role in Anomaly Detection} \\ 
\midrule
\endfirsthead

\multicolumn{4}{c}{\tablename\ \thetable{} -- continued from previous page} \\
\toprule
\textbf{Layer} & \textbf{Key Descriptors} & \textbf{Interpreted Specialization} & \textbf{Role in Anomaly Detection} \\ 
\midrule
\endhead

\multicolumn{4}{r}{\scriptsize Continued on next page} \\
\endfoot

\bottomrule
\endlastfoot

Layer 5 & Basic structural features & Basic Structure & Early structural feature extraction \\
\midrule
Layer 6 & photo, interior, bedroom, red, wildlife & Scene Type \& Basic Color & Scene context foundation \\
\midrule
Layer 7 & coffee, card, battery, interior, garden, pool, photography, video, size & Object Attributes \& Scene Context & Object-level feature encoding \\
\midrule
Layer 8 & photography, mesh, inspection, cell, draft, painting, frames, hat, football & Inspection-related \& Structural Patterns & Early inspection context awareness \\
\midrule
Layer 9 & cable, defense, category, tags, posters, size, digital, battery, utility, painting, profiles & Object Categories \& Measurement & Category and dimensional encoding \\
\midrule
Layer 10 & column, glass, yellow, silver, orange, pink, size, ford, dinner, bedroom, annotation, blue, red, green, laptops, determine & Color Spectrum \& Size Attributes & Chromatic and scale feature extraction \\
\midrule
Layer 11 & yellow, cleaning, menu, airport, font, bedroom, film, mesh, paint, auctions, slide, cable, directory, magazines, portal, flights, signs, instruments, kitchen, cisco, graphics, video, painting & Scene Diversity \& Functional Elements & Multi-context scene understanding \\
\midrule
Layer 12 & photography, poster, painting, video, photo, card, mesh, orange, buildings, advertisement, stories, size, lingerie, screen, glass, interior, wood, motorola, servers, compact, catalog, logo, chair & Visual Media \& Material Properties & Material-aware representations \\
\midrule
Layer 13 & posters, costa, flights, profiles, label, mirror, interior, blog, pool, sign, size, stories, video, cast, sub, hampshire, hat, orange, steps, application, flights, kansas, front, posters, counter, directors, line, film, signature, binding, kb, elementary, felt, sizes, photography, bath, black & Composition, Scale \& Material & Structural composition analysis \\
\midrule
Layer 14 & eye, codes, mount, lcd, glass, flag, column, tables, photography, regional, red, racing, elementary, houses, panel, appliances, table, pda, hat, film, extensive, blue, golden, micro, label, black, maintained, dental, errors, costa, chair, kong, shirts, sitemap, lighting & Detailed Inspection \& Display Attributes & Fine inspection feature encoding \\
\midrule
Layer 15 & sun, rain, interior, evening, bedroom, outdoors, cable, profiles, mouse, column, instruments, costa, poster, pocket, hat, victoria, guess, mouth, profiles, wildlife, adding, tours, lcd, reflect, flash, overall, wet, leather, painting, glass, tripadvisor, bath, goals, directions, nokia, pair, table, compensation, models, sign, hat, dress & Environmental Conditions \& Surface Properties & Lighting and surface awareness \\
\midrule
Layer 16 & Close-up of textured synthetic fur, Image with camouflage print, Glistening dew-covered foliage, Ephemeral soap bubble pattern, Intricate wood carving, Photo with cross-processing effect, Close-up of textured synthetic plastic, detailed reptile close-up, tribal tattoo-like designs, Close-up of textured synthetic wood, Artwork with stippling technique, ink wash painting style, Time-lapse trails, Minimalist architectural photography & Surface Texture \& Pattern Recognition & Foundational texture feature extraction \\
\midrule
Layer 17 & Aerial view, Point of view from below, Minimalist white backdrop, misty forest path, Rustic wooden textures, Earthy color tones, blue/green/purple color palette, Timeless black and white, sepia-toned vintage, Photo with cross-processing effect, gray color, Classic black and white cityscape, light leaks, Misty forest glade, Bustling cityscape at night, yellow color palette, Burst of colorful confetti, Ocean sunset silhouette, Artwork with stained glass window design, quadrilateral, Mondrian-like grids, Iconic landmarks, Intriguing forest pathway, Close-up of textured silk, Skyscrapers touching clouds, Minimalist white backdrop, Photograph taken indoors with low light, Colorful hot air balloons & Color Palette, Viewpoint \& Geometric Patterns & Mid-level color/structural encoding \\
\midrule
Layer 18 & size, trends, portal, novel, lcd, dog, ship, cup, overall, posters, portable, combined, shares, plate, newspaper, lake, cable, programme, bedroom, outdoors, airport, workshop, tables, glass, wall, restaurant, floor, cnet, papers, stream, controls, bookmark, signature, desk, instruments, accepted, thousands, atom, flowers, matches, interior, dish, legs, clinical, journal, ground, chair, nude, thailand, checks, models, inches, assembly, concern, vehicle, couple, cleaning, rear, vendor, bed, kid, relief, agreed, guides, specifications, cover, tennis & Workshop/Industrial Context \& Measurement & Industrial scene understanding \\
\midrule
Layer 19 & Artwork with intricate filigree patterns, Close-up of textured seashell, Bold geometric shapes, shattered glass sculptures, Vibrant watercolor painting, yellow color palette, Earthy color tones, cross-processing effect, Glowing neon cityscape, candlelit setting, Quaint countryside lane, Tranquil boating, Inviting bedroom atmosphere, Aerial view of agricultural field, A staircase, Bustling city intersection, Cozy living room ambiance, delicate soap bubble pattern, detailed insect close-up, camouflage print, Precise pocket watch, Close-up of textured synthetic fur, Celtic spiral designs, desert vista, Close-up of textured synthetic leather, A thread, cozy reading nook, Blossoming springtime blooms & Fine-grained Textures, Color Contrast \& Spatial Structure & High-level semantic texture \& scene understanding \\

\end{longtable}

\setlength{\tabcolsep}{6pt}
\renewcommand{\arraystretch}{1.0}

\end{document}